\documentclass{article}

 \usepackage[main, final]{neurips_2026}

\usepackage[utf8]{inputenc} 
\usepackage[T1]{fontenc}    
\usepackage{hyperref}       
\usepackage{url}            
\usepackage{booktabs}       
\usepackage{amsfonts}       
\usepackage{nicefrac}       
\usepackage{microtype}      
\usepackage{xcolor}         
\usepackage{graphicx}
\usepackage{amsmath}
\usepackage{amssymb}
\usepackage{cleveref}
\usepackage{multirow}   
\usepackage{algorithm}
\usepackage{algpseudocode}
\usepackage{listings}
\usepackage{zebra-goodies}
\usepackage{wrapfig}
\usepackage{caption}
\usepackage{subcaption}
\usepackage{bm}
\usepackage{xspace}
\usepackage{colortbl}
\usetikzlibrary{arrows.meta}

\definecolor{rankFirst}{rgb}{1.0, 0.70, 0.70}   
\definecolor{rankSecond}{rgb}{1.0, 0.85, 0.70}  
\definecolor{rankThird}{rgb}{1.0, 1.00, 0.70}   

\definecolor{tablethree}{rgb}{0.7, 1, 1}
\definecolor{tabletwo}{rgb}{0.7, 0.85, 1}
\definecolor{tableone}{rgb}{0.7, 0.7, 1}

\newcommand{\best}{\cellcolor{rankFirst}}
\newcommand{\sbest}{\cellcolor{rankSecond}}

\makeatletter
\DeclareRobustCommand\onedot{\futurelet\@let@token\@onedot}
\def\@onedot{\ifx\@let@token.\else.\null\fi\xspace}

\def\eg{\emph{e.g}\onedot} 
\def\ie{\emph{i.e}\onedot}

\def\etal{\emph{et al}\onedot}
\makeatother

\def\fwd{\mathrm{fwd}}
\def\bwd{\mathrm{bwd}}
\def\loss{\mathcal{L}}
\def\sg{\mathrm{sg}}

\def\glb{g}
\def\lcl{l}
\def\CLS{\texttt{[CLS]}}
\newcommand{\real}{\mathbb{R}}
\definecolor{oursAccent}{RGB}{55, 140, 185}

\title{BAL: Bidirectional Autoregression in Latent Space for Learning Human Movement Representations}

\author{%
  Genki Kinoshita \qquad\qquad\qquad\qquad\qquad
  Ko Nishino\\[0.8em]
  Graduate School of Informatics, Kyoto University\\[0.6em]
  \texttt{gkinoshita@vision.ist.i.kyoto-u.ac.jp} \\[0.3em]
  \textcolor{oursAccent}{\url{https://github.com/kyotovision-public/bal}}
}

\begin{document}

\maketitle

\begin{abstract}
Learning semantically meaningful human movement representations from 3D pose sequences is essential for human behavior modeling.
Existing self-supervised methods derive supervision from artificial assumptions, \eg, invariance to augmentation within a certain range or recoverability of masked regions at a certain temporal scale, introducing heuristic hyperparameters that make the learned semantics sensitive to their settings.
We propose BAL, a self-supervised framework that instead exploits supervision intrinsic to human movement: the natural ordering structure of everyday behavior.
BAL autoregressively predicts in a semantically abstract latent space, analogous to how humans anticipate others' actions rather than their precise joint configurations.
Bidirectional autoregressive prediction, both forward and backward in time, fully leverages these ordering constraints while enabling representation extraction using both past and future context at inference.
To suppress representation collapse, BAL decomposes each prediction target into a global component capturing sequence-level semantics and a local component capturing context-independent movement.
We prove that the local targets retain nontrivial angular diversity unless the representations are completely collapsed, preserving a meaningful training signal.
Experiments on large-scale datasets demonstrate that BAL's representations are highly effective for motion captioning, forecasting, and interpolation, outperforming baselines on most evaluation metrics.
\end{abstract}

\section{Introduction}
Understanding human movement semantics underpins a broad range of behavior modeling tasks, from motion captioning to forecasting.
Among available modalities for movement understanding, 3D pose sequences are particularly attractive: they abstract away appearance and scene context, are inherently privacy-preserving, and are available at a rapidly growing scale. Ideally, movement semantics would be learned from this abundant data using only self-supervision.
Yet learning representations that capture these semantics from 3D pose sequences alone remains an open challenge.

Existing self-supervised methods share a fundamental limitation: the supervisory signal is defined by artificial design choices.
Contrastive methods \cite{CrosSCLR,HiCLR,USDRL,AimCLR} require representations to be invariant under hand-crafted augmentations with chosen strengths; masked methods \cite{MotionBERT,BehaveMAE,MacDiff,MAMP,A4Mer,SkeletonMAE} require masked regions to be recoverable from context at a chosen temporal scale.
The resulting representations are sensitive to these choices, and obtaining semantics suited to a target task often requires dataset- or task-specific hyperparameter tuning.
This sensitivity is not incidental; it follows directly from the fact that the supervision is externally imposed rather than grounded in properties of the data itself.

A more principled alternative is to exploit supervision intrinsic to human movement.
Daily behavior exhibits a consistent ordering structure toward goal completion: one pulls out a chair before sitting, opens a refrigerator before reaching inside, and turns a door handle before walking through.
These actions are not independent---they unfold as temporally dependent steps within goal-directed processes.
This structure is a natural source of supervision: if representations capture movement semantics, they should reflect these ordering constraints.

The most direct way to exploit this ordering structure is to predict the future from the past.
However, frame-level autoregressive prediction in pose space does not naturally produce semantic representations.
Short-term joint trajectory smoothness is sufficient to minimize such a loss, so representations tend to encode local pose kinematics rather than action-level semantics.
Predicting farther into the future does not resolve this, because the same action can manifest as different poses depending on the person, context, and style; pose-space prediction targets are inherently ambiguous, pushing the model to average over plausible futures rather than distinguish their underlying semantics.

Humans do not reason about others' behavior this way.
Rather than estimating future joint positions, we anticipate behavior in semantically abstract units---sitting down, taking something out, leaving the room.
Moreover, this semantic prediction operates not only from past to future but also from future to past: knowing someone is reaching into a refrigerator, we naturally infer they opened it moments before.
We propose BAL (\textbf{B}idirectional \textbf{A}utoregression in \textbf{L}atent space), a self-supervised framework that turns these observations into a learning objective.
BAL divides a pose sequence into fixed-length temporal chunks, compresses each chunk into a single latent token, and trains by predicting these tokens autoregressively in both temporal directions, rather than predicting raw poses.
Since latent tokens serve as semantic abstractions of chunk-level movements, ordering constraints among them provide a self-supervisory signal grounded in the structure of behavior itself, reducing the reliance on heuristic hyperparameters to define it.
During training, bidirectional prediction more fully exploits the ordering constraints as supervision. Moreover, at inference, it enables the representation extractor to process all chunks without causal masking, extracting chunk representations based on both past and future context in a single forward pass.

A key challenge is that the prediction targets must be learned jointly with their predictors, creating a risk of representation collapse---the extractor degenerates to nearly constant outputs, making prediction trivially easy but semantically vacuous.
We address this by decomposing each prediction target into a global component (the mean across all latent tokens in the sequence) and a local component (the residual from that mean).
We prove that the L2-normalized local targets always retain nontrivial angular diversity (their maximum pairwise angle is at least $90^{\circ}$) as long as the latent tokens are not all identical.
This means that even as the extractor starts collapsing, predicting normalized local targets does not become trivially easy, helping to preserve a gradient signal that resists further collapse.

We evaluate BAL on large-scale daily activity datasets across three downstream tasks: motion captioning, long-term motion forecasting, and long-term motion interpolation.
BAL outperforms the compared self-supervised methods, including masked modeling and contrastive approaches, on most evaluation metrics, suggesting that ordering constraints in latent space provide a more effective supervisory signal for learning movement semantics.

\section{Related work}

\paragraph{Self-supervised representation learning from 3D poses.}
A variety of self-supervised objectives have been proposed to learn movement representations from 3D pose sequences \cite{MS2L,MotionConsistencyAndContinuity,ActCLR,SkeletonMAE,MAMP,PCM3,PUMPS,S-JEPA,BehaveMAE,USDRL,MacDiff,MotionBERT,HiCLR,GFP}.
They are broadly grouped into masked modeling \cite{MotionBERT,MacDiff,PUMPS} and contrastive learning \cite{ActCLR,USDRL}.
The former learns by predicting masked input regions, whereas the latter learns by making representations invariant to data augmentations.
In both cases, supervision depends on artificial design choices, such as mask size or augmentation type and strength.
These choices affect the content of the learned representations, yet are difficult to select and often require task- or dataset-specific tuning.
In contrast, BAL uses the ordering structure in daily human behavior as supervision, reducing the reliance on such heuristic design choices.

Most masked-modeling methods \cite{MotionBERT,MacDiff,BehaveMAE} compute losses in observation space.
Their training can therefore be dominated by minimizing fine-grained joint differences.
BAL instead performs autoregressive prediction in latent space, shifting the focus from reproducing minor pose variations to capturing underlying movement semantics.

A4Mer \cite{A4Mer} also defines its objective in latent space using JEPA \cite{JEPA}.
However, it requires a complex two-stage training and inference pipeline, along with heuristic hyperparameters, to learn representations at semantically meaningful units.
While this is a reasonable direction, we argue that the more critical issue is the choice of self-supervisory signal.
Rather than discovering such units with a complex framework, BAL uses the ordering constraints intrinsic to human movement as supervision, enabling simpler, more scalable learning with fewer hyperparameters to define supervision while producing more effective representations.

\paragraph{Self-supervised learning with autoregressive prediction} has been successful in various fields, including text \cite{ELMo,GPT}, speech \cite{CPC,wav2vec}, images \cite{PixelCNN,ImageTransformer,PixelRNN}, and videos \cite{VideoGPT,TATS}, while it remains underexplored for human movement.
In vision, iGPT \cite{iGPT} and AIM \cite{AIM} show that representations learned with autoregression achieve higher linear-probe classification accuracy than those learned with masked modeling, suggesting the potential of autoregressive objectives for representation learning.

Autoregressive methods differ in whether they predict in the observation space or in a latent space.
A representative example of the latter is CPC \cite{CPC}, which predicts future latent representations from an autoregressively computed context. DPC \cite{DPC} and wav2vec \cite{wav2vec} adapt this idea to RGB videos and speech recognition, respectively.
These methods, however, rely on contrastive losses, which require complex objective design, including negative sampling.
More recent methods instead directly regress latent representations, as in NEPA \cite{NEPA} for image pretraining and LeWM \cite{LeWM} for world models.
BAL follows this direction by minimizing the distance between predicted and target representations.

Still, existing regression-based latent autoregressive methods have notable limitations.
NEPA predicts shallow features produced by a single convolutional layer, which remain tied to local input details and lack semantic abstraction; indeed, it reports weak linear-probe performance.
LeWM predicts deeper Transformer representations, but requires SIGReg \cite{SIGReg} for stable training, introducing additional hyperparameters.
Moreover, we find that SIGReg is not sufficient to suppress representation collapse on human movement data.
Like LeWM, BAL predicts deep latent representations, but it mitigates representation collapse through a simple decomposition of prediction targets into global and local movement components, without introducing hyperparameters.

\paragraph{Bidirectional autoregression.}
Since autoregressive prediction is inherently directional, an important challenge is how to combine autoregressive self-supervision with bidirectional contextual inference.
In text and speech, several works address this by training separate past-to-future and future-to-past models \cite{TagLM,ELMo,BI-APC}.
Although this exploits both temporal directions, it requires computing two direction-specific representations at inference time, increasing computational cost.

This design largely stems from coupling representation extraction and autoregressive prediction in a single model.
Since past-to-future and future-to-past predictions follow different conditional distributions, modeling both directions naturally leads to direction-specific models.
BAL instead decouples these two roles by using a shared representation extractor with two direction-specific predictors.
The predictors model the direction-specific conditional distributions required for autoregressive prediction, allowing the shared extractor to focus on representing the movement in each corresponding chunk rather than encoding direction-specific predictive states.
By using the same extractor to learn past-conditioned and future-conditioned views of each chunk, BAL places these representations in a shared latent space.
At inference time, this gives BAL a way to extract representations without causal masks while considering bidirectional context in a single forward pass.

\section{BAL}
\label{sec:method}

\begin{figure}
  \centering
  \includegraphics[width=\linewidth]{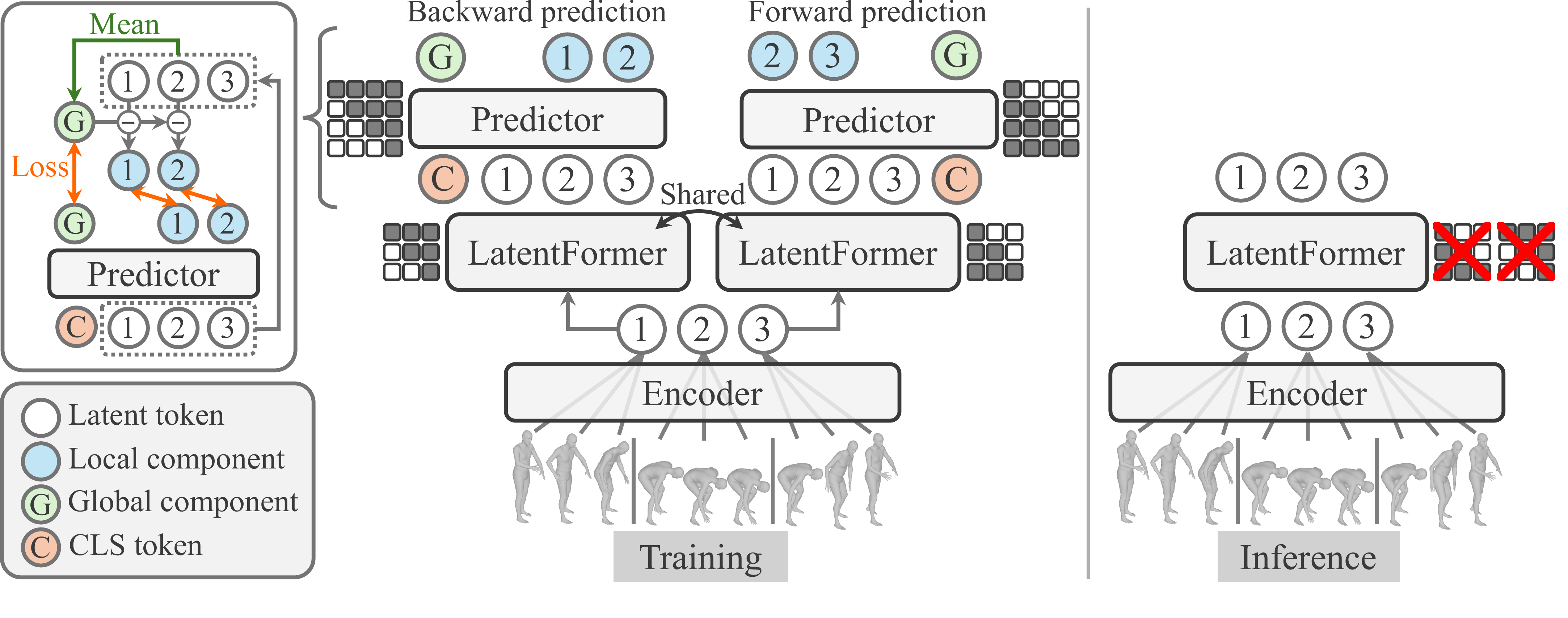}
  \caption{
BAL learns movement representations from 3D pose sequences by bidirectional latent-space autoregressive prediction.
The Encoder consolidates each chunk into a latent token, and a shared LatentFormer processes these tokens in both directions using causal and reversed causal masks.
Two direction-specific predictors predict local components of adjacent latent tokens, while the global component is predicted from a learnable \CLS~token.
At inference time, the LatentFormer extracts chunk representations without causal masking, enabling bidirectional contextual inference.
  }
  \label{fig:overview}
\end{figure}

Our goal is to learn movement representations from 3D pose sequences by leveraging the natural ordering constraints inherent in daily human behavior, with minimal dependence on heuristic hyperparameters for defining the supervisory signal.
We provide an overview of BAL in Fig.~\ref{fig:overview}.

\subsection{Learning by autoregression in latent space}

BAL is trained by autoregressively predicting latent tokens.
Each token represents a temporal chunk of movement, reflecting that the ordering structure in daily behavior emerges more naturally between temporally extended units than between individual frames.
By predicting chunk-level semantic abstractions rather than fine-grained poses, BAL uses behavioral ordering as semantic supervision.

BAL consists of a representation extractor and a predictor.
As in A4Mer \cite{A4Mer}, the representation extractor comprises an Encoder $f_E$ and a LatentFormer $f_L$.
Given a 3D pose sequence $X$, BAL divides it into $M$ temporal chunks $X_{1:M}$.
For each chunk $X_m$, the Encoder initializes a chunk embedding by max pooling over the frames in $X_m$ and updates it by alternating self-attention within $X_m$ and cross-attention from the chunk embedding to the frames in $X_m$:
\begin{equation}
    h_m = f_E(X_m)\,,
\end{equation}
where $h_m$ denotes the resulting chunk embedding.
The LatentFormer, a Transformer Encoder \cite{Transformer}, models temporal dependencies across chunks and yields a latent token $z^\fwd_m$ for each chunk, while applying a causal mask to prevent access to future chunks during autoregressive learning:
\begin{equation}
    z^\fwd_m = f_L(h_{\le m})\,.
\end{equation}
Since self-attention in the Encoder is confined to each chunk, the Encoder is effectively limited to local aggregation into chunk embeddings.
The LatentFormer therefore operates on already abstracted chunk representations, allowing it to reason about inter-chunk relationships at a semantic level.

The predictor $g^\fwd$, a Transformer Encoder with causal masking, then takes $z^\fwd_{\le m}$ and predicts the next latent token:
\begin{equation}
    \hat{z}^\fwd_{m+1} = g^\fwd(z_{\le m}^\fwd)\,.
\end{equation}
We apply normalization to the prediction and the target as appropriate for the chosen objective (see Sec.~\ref{subsec:supp:arch}), and minimize the distance between them:
\begin{equation}
\label{eq:fwd}
    \loss^\fwd = \frac{1}{M-1} \sum_{m=1}^{M-1} D\left(\hat{z}^\fwd_{m+1}, \sg\left(z^\fwd_{m+1}\right) \right)\,,
\end{equation}
where $D(\cdot,\cdot)$ denotes a distance function that incorporates normalization (\eg, L2 normalization or layer normalization \cite{LayerNorm}), and $\sg(\cdot)$ denotes stop-gradient.

\subsection{Bidirectional autoregression and inference}

\label{subsec:Bidirectional}
To fully exploit natural ordering constraints as self-supervision while allowing inference that considers both past and future context, BAL learns by latent-space autoregressive prediction in both temporal directions.
The forward and backward directions share the representation extractor but use separate predictors: the LatentFormer applies a standard causal mask in the forward direction and a reversed causal mask in the backward direction, while a dedicated backward predictor $g^\bwd$ is introduced because past- and future-conditioned predictions follow different conditional distributions.
With reversed causal masks in both the LatentFormer and the predictor, the backward direction yields
\begin{equation}
    z_m^\bwd = f_L(h_{\ge m}), \qquad
    \hat{z}_{m-1}^\bwd = g^\bwd(z_{\ge m}^\bwd)\,.
\end{equation}
The backward loss is defined analogously to Eq.~(\ref{eq:fwd}):
\begin{equation}
\label{eq:bwd}
    \loss^\bwd
    =
    \frac{1}{M-1}
    \sum_{m=2}^{M}
    D\left(
        \hat{z}_{m-1}^\bwd,
        \sg\!\left(z^\bwd_{m-1}\right)
    \right)\,.
\end{equation}
By sharing the representation extractor and localizing the asymmetry between the two conditional distributions to the predictors, BAL reduces the need for the extracted representations themselves to encode direction-specific predictive states.
The shared extractor instead learns past-conditioned and future-conditioned views of each chunk in a common latent space, which gives us a way to remove the causal masks at inference time.
As a result, the extractor can produce representations while taking both past and future context into account in a single forward pass. We show in Sec.~\ref{subsec:ablation} that this bidirectional inference improves downstream performance.

\subsection{Decomposition into local and global components}
\label{subsec:decompose}

In practice, we find that directly minimizing Eqs.~(\ref{eq:fwd}) and (\ref{eq:bwd}) pushes the representations toward collapse, where the extractor outputs highly similar representations.
This arises from a self-reinforcing loop: once the extractor starts producing similar representations, the predictors can reduce the loss simply by outputting near-constant predictions regardless of their inputs, which further drives training toward collapse.
Conversely, if predictions from such similar representations remain difficult and incur a sufficiently large loss, collapse can be suppressed.
This suggests that preserving sufficient diversity in the prediction targets can suppress collapse.

Based on this, BAL decomposes each target representation into a global component that captures the semantics of the entire sequence and a local component that captures the movement of each chunk independent of the sequence-level semantics.
Given latent tokens $z_{1:M}$ of either direction,
we define the global component $z^\glb$ as their mean and the local component $z^\lcl_m$ as the residual:
\begin{equation}
    z^\glb = \frac{1}{M} \sum_{m=1}^{M} z_m\,,\qquad z_m^\lcl = z_m - z^\glb\,.
\end{equation}

A key property of this decomposition is that the L2-normalized local components $u_m=z^\lcl_m/\|z^\lcl_m\|$ always retain nontrivial diversity when all local components are nonzero.
Formally, writing $a_m=\|z^\lcl_m\|>0$ and $\mu=\min_{m\neq n}u_m^\top u_n$, the identity $\sum_{m=1}^M z^\lcl_m=0$ gives $\sum_{m=1}^M a_m u_m=0$.
Taking the squared norm yields
\begin{equation}
    0=\left\|\sum_{m=1}^M a_m u_m\right\|^2
    =\sum_{m=1}^M a_m^2 + 2\sum_{m<n}a_m a_n\,u_m^\top u_n
    \ge
    \sum_{m=1}^M a_m^2 + 2\mu\sum_{m<n}a_m a_n\,,
\end{equation}
and therefore
\begin{equation}
    \mu \le -\frac{\sum_{m=1}^M a_m^2}{2\sum_{m<n}a_m a_n}\,.
\end{equation}
Moreover, Cauchy--Schwarz implies
\begin{equation}
    \left(\sum_{m=1}^M a_m\right)^2\le M\sum_{m=1}^M a_m^2 \Longleftrightarrow 2\sum_{m<n}a_m a_n \le (M-1)\sum_{m=1}^M a_m^2\,.
\end{equation}
Combining these inequalities gives $\mu\le -1/(M-1)$. Since $\angle(u_m,u_n)=\arccos(u_m^\top u_n)$ and $\arccos$ is decreasing, we obtain
\begin{equation}
    \max_{m\neq n}\angle(u_m,u_n)
    =
    \arccos\!\left(\min_{m\neq n}u_m^\top u_n\right)
    \ge
    \arccos\!\left(-\frac{1}{M-1}\right)
    \ge \frac{\pi}{2}\,.
\end{equation}
Thus, when all local components are nonzero, at least one pair of normalized local components is separated by an angle of at least $90^{\circ}$.
Hence, even when the extractor produces nearly identical representations, the normalized local components remain nontrivially diverse, making them difficult for the predictors to infer from such similar representations; penalizing this prediction error should therefore discourage such degeneracy and suppress collapse.
An analogous argument holds when layer normalization is used instead of L2 normalization (see Appendix~\ref{subsec:supp:decompose}).

The predictors in BAL predict the local and global components separately.
The forward predictor $g^\fwd$ predicts $z_{m+1}^{\lcl}$ from $z_{\le m}^{\fwd}$, while the backward predictor $g^\bwd$ predicts $z_{m-1}^{\lcl}$ from $z_{\ge m}^{\bwd}$.
In each direction, the global component is predicted from a learnable \CLS~token that attends to the entire sequence without causal masking to summarize the global movement but is not attended to by other tokens.
Let $\hat{z}_{m+1}^{\lcl,\fwd}$ and $\hat{z}_{m-1}^{\lcl,\bwd}$ denote the predicted local components in the forward and backward directions, and let $\hat{z}^{\glb,\fwd}$ and $\hat{z}^{\glb,\bwd}$ denote the global components predicted from the \CLS~token.
Using this notation, the local and global losses can be written as
\begin{align}
    \label{eq:local}
    \loss^{\lcl,\fwd}_m
    =
    D\left(
    \hat{z}_{m+1}^{\lcl,\fwd},
    \mathrm{sg}\left(z_{m+1}^\lcl\right)
    \right), \,&\,
    \loss^{\lcl,\bwd}_m
    =
    D\left(
    \hat{z}_{m-1}^{\lcl,\bwd},
    \mathrm{sg}\left(z_{m-1}^\lcl\right)
    \right)\,, \\
    \loss^{\glb,\fwd}
    =
    D\left(
    \hat{z}^{\glb,\fwd},
    \mathrm{sg}\left(z^\glb\right)
    \right), \,&\,
    \loss^{\glb,\bwd}
    =
    D\left(
    \hat{z}^{\glb,\bwd},
    \mathrm{sg}\left(z^\glb\right)
    \right)\,.
    \label{eq:global}
\end{align}
The final objective is
\begin{equation}
\label{eq:total}
    \loss
    =
    \frac{1}{M-1}\sum_{m=1}^{M-1} \left( \loss^{\lcl,\fwd}_m + \lambda^\fwd_m \loss^{\glb,\fwd} \right)
    +
    \frac{1}{M-1}\sum_{m=2}^{M} \left( \loss^{\lcl,\bwd}_m + \lambda^\bwd_m \loss^{\glb,\bwd} \right) \,,
\end{equation}
where $\lambda^\fwd_m = \sg(\loss^{\lcl,\fwd}_m/\loss^{\glb,\fwd})$ and $\lambda^\bwd_m = \sg(\loss^{\lcl,\bwd}_m/\loss^{\glb,\bwd})$ dynamically balance the magnitudes of the local and global losses so that neither term dominates optimization.

A4Mer, which uses JEPA rather than latent-space autoregressive prediction, also decomposes representations into global and local components, but with a different motivation: there, the decomposition serves to learn reusable representations that are independent of context.
We reinterpret it as a principled way to suppress representation collapse and provide a theoretical account of how it does so.
Moreover, whereas A4Mer decomposes the predicted latent tokens when computing the loss, the predictors in BAL directly predict the global and local components.
This clearer separation of the prediction targets makes it easier for the predictors to learn what each output should represent.
We empirically find that this leads to better downstream performance than the A4Mer-style variant.

\section{Experiments}
\label{sec:experiments}

We evaluate the effectiveness of BAL's representations on three downstream tasks that require understanding movement semantics: motion captioning, forecasting, and interpolation.
As baselines, we compare BAL with self-supervised representation learning methods based on different pretext tasks.
Each baseline follows its official implementation, including its input pose representation, network architecture, and training hyperparameters.
We further conduct a qualitative motion retrieval analysis to inspect whether the learned representation space reflects semantic similarity between motions.

When training task-specific heads for downstream tasks, we freeze the representation extractors for all tasks, since finetuning them could make performance depend more heavily on model size and network architecture than on the quality of the representations learned through the self-supervised objective of each method.
We provide architectural details and training settings in Appendix~\ref{sec:supp:impldetail}.

\subsection{Datasets}

BAL leverages the natural ordering structure in everyday human behavior as self-supervision.
We therefore pretrain both BAL and the baselines on a mixture of three datasets that contain natural daily activities: AMD \cite{A4Mer}, Embody 3D \cite{Embody3D}, and Nymeria \cite{Nymeria}.

AMD contains about 14.2 hours of daily household activities recorded in a real house.
We follow the official protocol and split the data into training, validation, and test sets with an 8:1:1 ratio.

Nymeria contains approximately 300 hours of daily activities recorded in indoor and outdoor environments, including multi-person interactions.
We separate sessions by whether they have text annotations so that annotated sessions are included in every split for motion captioning.
For each subset, we hold out 20\% of the sessions as the test set and split the remaining sessions into training and validation sets at an 8:2 ratio.

Embody 3D (E3D) contains about 500 hours of indoor motion sequences.
While recorded in a smaller and more constrained environment than the others, it covers diverse behaviors such as charades and multi-person conversations.
We split E3D in the same way as Nymeria.

Following A4Mer~\cite{A4Mer}, we downsample input sequences to 5 fps to efficiently handle long sequences while preserving movement semantics.
For downstream tasks, we train the forecasting and interpolation heads on the same mixture and evaluate them on each dataset, and train and evaluate the captioning head on E3D and Nymeria, which provide text annotations.

In both pretraining and downstream tasks, the models take input sequences of up to 30 seconds.
BAL uses a fixed chunk length of 4 seconds, following the definition of atomic actions in Nymeria as movements lasting 3--5 seconds.

\begin{table*}[t!]
    \centering
    \scriptsize
    \caption{Comparison of methods pretrained on AMD, Nymeria (Nym), and Embody 3D (E3D), for motion captioning, forecasting, and interpolation. R-L and BERT denote ROUGE-L and BERTScore, respectively. BAL achieves the best performance on most metrics across tasks.}
    \label{tab:results}
    \renewcommand{\arraystretch}{1.2}
    \setlength{\tabcolsep}{1.8pt}
    \begin{tabular}{@{}ll|cccccccc|ccc|ccc@{}}
    \toprule
    \multirow{3}{*}{\textbf{Method}} & \multirow{3}{*}{\textbf{Objective}} & \multicolumn{8}{c|}{\textbf{Captioning} $\uparrow$} & \multicolumn{3}{c|}{\multirow{2}{*}{\textbf{Forecasting} (mm) $\downarrow$}} & \multicolumn{3}{c}{\multirow{2}{*}{\textbf{Interpolation} (mm) $\downarrow$}} \\
    & & \multicolumn{2}{c}{BLEU-1~\cite{BLEU}} & \multicolumn{2}{c}{R-L~\cite{ROUGE}} & \multicolumn{2}{c}{CIDEr~\cite{CIDEr}} & \multicolumn{2}{c|}{BERT~\cite{BERTScore}} & \multicolumn{3}{c|}{} & \multicolumn{3}{c}{} \\
    & & Nym & E3D & Nym & E3D & Nym & E3D & Nym & E3D & AMD & Nym & E3D & AMD & Nym & E3D \\
    \midrule
    A4Mer \cite{A4Mer}           & JEPA              & 15.7        & 32.9        & 20.5        & 22.1        &  4.3        &  4.3        &  8.7        & 10.2        & 641.6        & \sbest 359.0 & 381.3       & 163.7        & 145.1        &  86.3 \\
    PUMPS \cite{PUMPS}           & Reconstruct       &  8.2        & 58.7        & 17.0        & 44.8 & 16.2        & 17.3        &  9.5        & 34.3        & 186.4        & 453.6        & 111.7       & 195.6        & 472.4        &  89.2 \\
    USDRL \cite{USDRL}           & Contrastive       & 27.1        & 58.0        & 34.8        & \best 45.3  & 32.5        & \sbest 18.4 & 17.7        & \sbest 35.0 & \sbest 136.1 & 373.8        &  90.2       & 117.7        & 424.6        & \sbest 59.2 \\
    BehaveMAE \cite{BehaveMAE}   & MAE               & 33.9        & \sbest 58.8 & 38.3        & 44.6        & 49.4        & 17.4        & 24.6        & 33.6        & 195.5        & 394.8        & 152.7       & 135.6        & \sbest 103.5 &  77.8 \\
    MotionBERT \cite{MotionBERT} & Mask \& 2D$\to$3D & 24.3        & 58.4        & 36.6        & \sbest 45.2        & 19.9        & 16.7        & 15.8        & \best 35.2  & 976.6        & 1623.2       & 248.7       & 136.7        & 348.7        &  69.5 \\
    MacDiff \cite{MacDiff}       & Mask \& Denoise   & \sbest 36.7 & 58.2        & \best 41.4  & 44.7        & \best 65.6  & 17.8        & \sbest 28.5 & 33.6        & 160.0        & 371.8        & \sbest 73.2 & \best 112.7  & 454.1        & \best 56.6 \\
    \midrule
    \textbf{BAL}                 & Latent AR         & \best 37.3  & \best 58.9  & \sbest 41.2 & \best 45.3  & \sbest 63.1 & \best 18.6  & \best 29.6  & \sbest 35.0 & \best 90.5   & \best 82.1   & \best 63.8  & \sbest 117.5 &  \best 77.0 & \sbest 59.2  \\
    \bottomrule
    \end{tabular}
\end{table*}

\subsection{Motion captioning}
\label{subsec:captioning}

To examine whether BAL's representations capture semantics that can be translated into natural language, we perform motion captioning with a large language model (LLM).
We project the extracted representations into the LLM text-embedding space using a two-layer MLP, and feed the resulting embeddings into the LLM together with fixed system and user prompts to generate captions.

We use Llama 3.1 8B Instruct \cite{Llama} as the LLM and keep it frozen, training only the projection MLP to measure pure semantic awareness of the learned representations.
Since E3D and Nymeria differ in annotation granularity, we train a separate projection MLP for each dataset.

\begin{figure}[t]
  \centering
  \includegraphics[width=\linewidth]{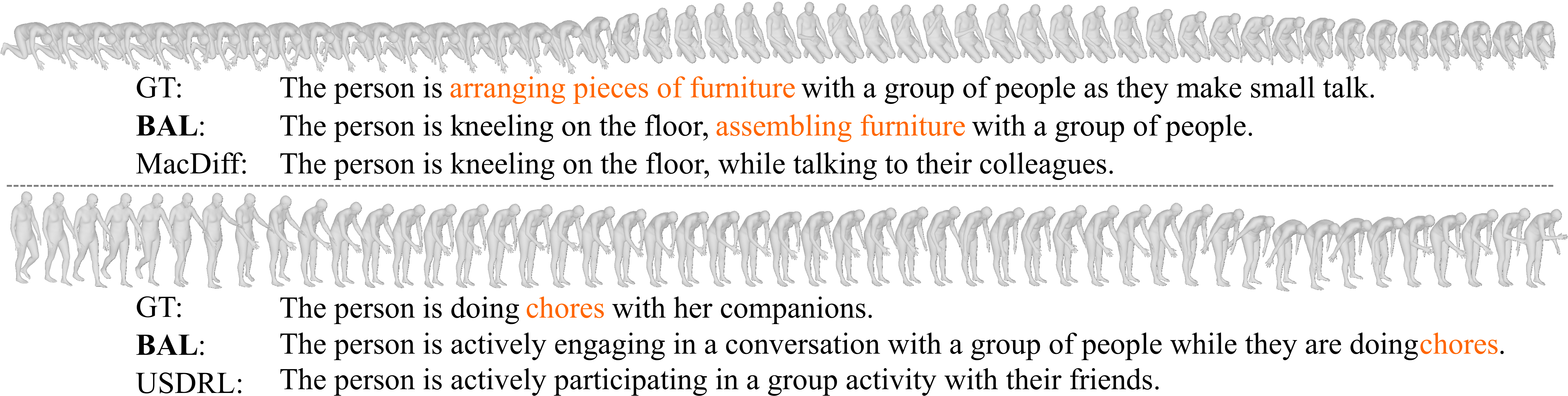}
  \caption{
  Motion captioning results. BAL generates more accurate descriptions than the baselines, suggesting that its representations capture the semantics of the movement more effectively.
  }
  \label{fig:captioning}
\end{figure}

Tab.~\ref{tab:results} reports results on standard text generation metrics.
BAL achieves higher scores than the baselines on most metrics. This suggests that its representations better capture movement semantics, allowing the LLM to describe the movement more accurately, as illustrated in Fig.~\ref{fig:captioning}.

\subsection{Long-term motion forecasting}
\label{subsec:forecast}

If the learned representations capture movement semantics, predicting future representations in latent space should enable forecasting at the semantic level rather than mere smooth extrapolation of poses.
To realize such semantically informed motion forecasting, we follow A4Mer and introduce a forecasting head that autoregressively predicts the next latent token from the latent token sequence extracted by the representation extractor.
The predicted latent tokens are then mapped back to pose space by a decoder, which is trained independently of the forecasting head.
We forecast poses up to 4 seconds ahead and evaluate using MPJPE averaged over all predicted frames.

Tab.~\ref{tab:results} shows the results.
BAL consistently achieves lower errors than the baselines, suggesting that its representations are more semantically informative and enable the forecasting head to reason about plausible future movement. This is also reflected in the qualitative results in Fig.~\ref{fig:forecast}.

\subsection{Long-term motion interpolation}
\label{subsec:interp}

Representations that capture semantics should also help infer missing motion between observed past and future poses.
To test this, we follow A4Mer and introduce an interpolation head that predicts latent tokens for the missing interval from those of the observed poses.
As in forecasting, the predicted tokens are decoded into poses with the decoder.

For training and evaluation, we randomly mask a 4- to 12-second interval in each input sequence and interpolate the motion within the masked region.
We evaluate the interpolated poses using MPJPE averaged over all masked frames.

\begin{figure}[t!]
    \centering
    \begin{subfigure}[b]{0.34\textwidth}
        \centering
        \includegraphics[width=\linewidth,trim=1pt 3pt 3pt 1pt,clip]{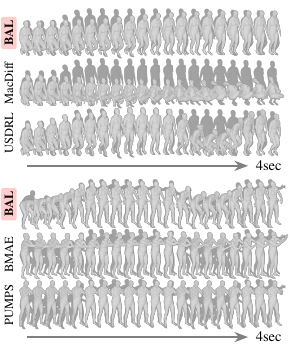}
        \caption{}
        \label{fig:forecast}
    \end{subfigure}
    \hfill
    \begin{subfigure}[b]{0.63\textwidth}
        \centering
        \includegraphics[width=\linewidth,trim=1pt 3pt 3pt 1pt,clip]{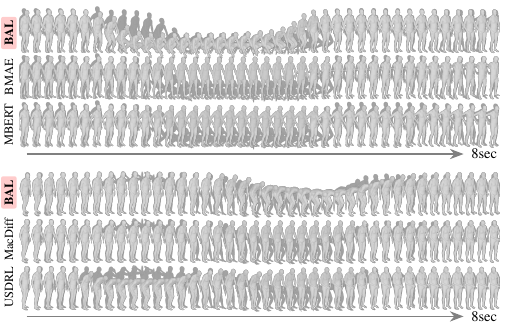}
        \caption{}
        \label{fig:interp}
    \end{subfigure}
    \caption{
    (a) Forecasted poses through autoregressive latent token prediction and (b) interpolated poses with latent token interpolation. The dark-colored SMPL~\cite{SMPL} meshes denote the ground truth. BMAE and MBERT denote BehaveMAE and MotionBERT, respectively. The representations learned by BAL enable more accurate and semantically plausible forecasting and interpolation.
    }
\end{figure}

Tab.~\ref{tab:results} shows that BAL achieves lower interpolation errors than most of the baselines, and Fig.~\ref{fig:interp} provides qualitative examples.
These results suggest that BAL's representations enable the interpolation head to infer semantically plausible missing motion from surrounding observations.

In Appendix~\ref{subsec:supp:largeError}, we further analyze BAL's large gains in motion forecasting and interpolation by aligning the global translation with the ground truth. This reduces the baselines' errors, suggesting that weaker encoding of global translation partly explains their performance gap with BAL. Yet BAL still outperforms the baselines on most metrics after the alignment.

\subsection{Motion retrieval}

\begin{figure}[t]
    \centering
    \begin{minipage}[t]{0.66\linewidth}
        \vspace{0pt}
        \centering
        \includegraphics[width=\linewidth]{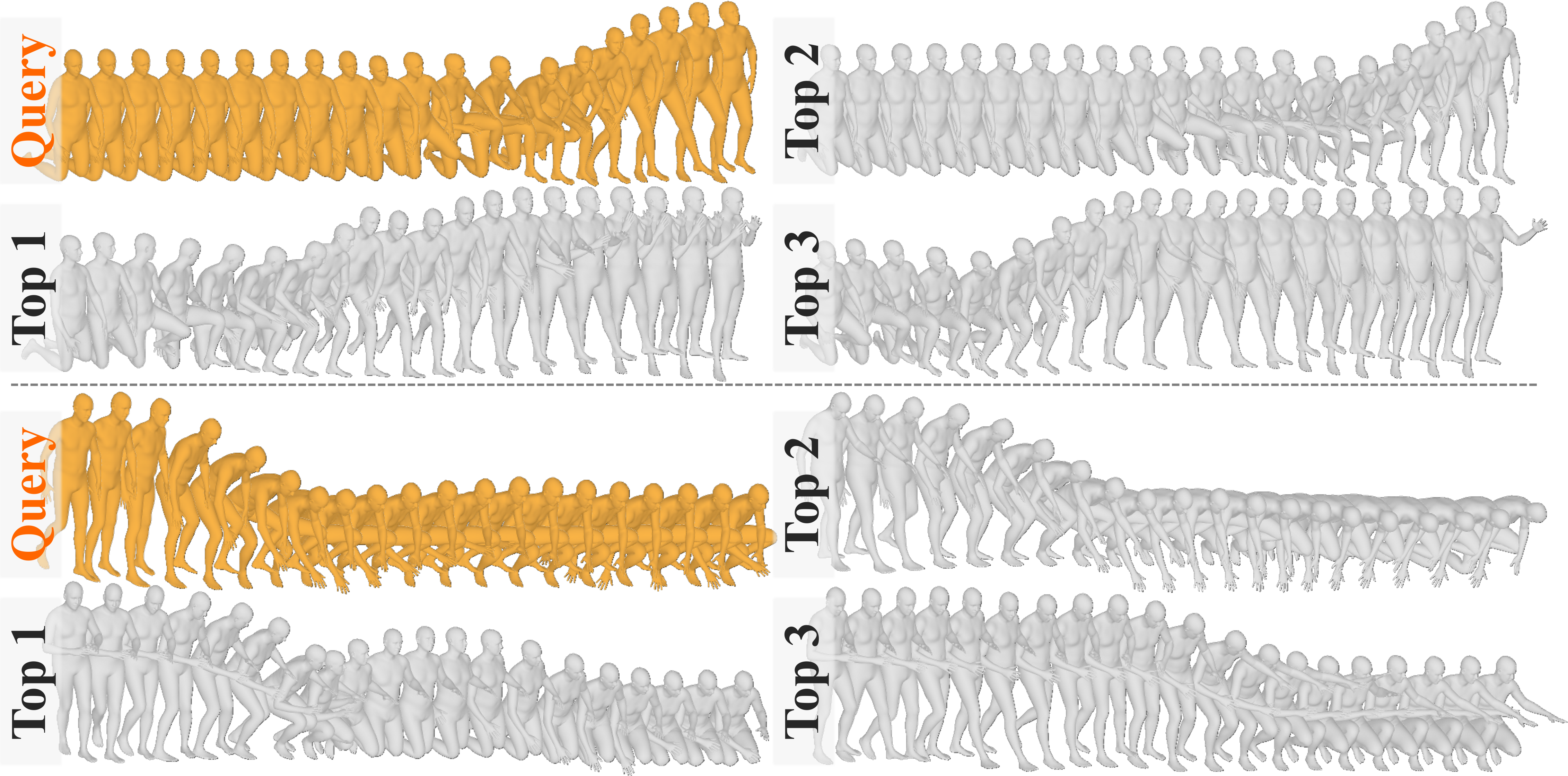}
        \captionof{figure}{
            Motion retrieval results. The three nearest neighbors to each query in the representation space correspond to semantically similar movements.
        }
        \label{fig:retrieval}
    \end{minipage}
    \hfill
    \begin{minipage}[t]{0.32\linewidth}
        \vspace{0pt}
        \centering
        \includegraphics[width=\linewidth]{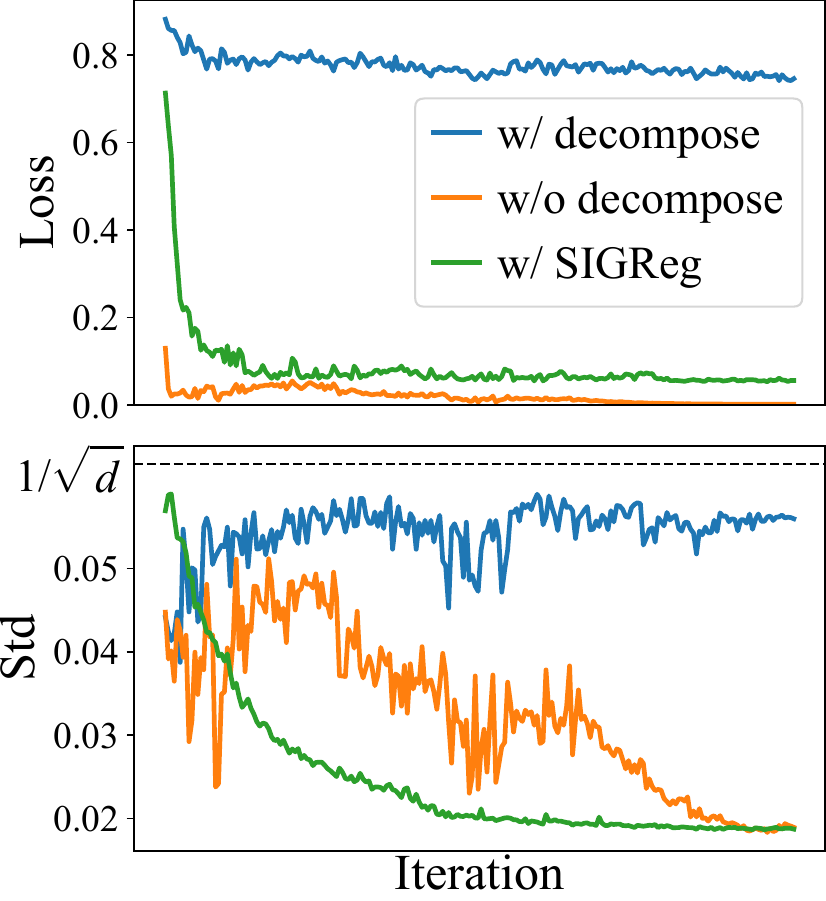}
        \vspace{-15pt}
        \captionof{figure}{
            Training loss and feature standard deviation during pretraining.
        }
        \label{fig:loss_std}
    \end{minipage}
\end{figure}

We perform motion retrieval on the pooled test splits of the three datasets to qualitatively analyze the learned representation space.
For each query motion, we retrieve the three nearest neighbors in BAL's representation space, excluding the query itself.
As shown in Fig.~\ref{fig:retrieval}, the retrieved motions are semantically similar to the queries even when they come from different datasets, suggesting that the learned representation space reflects semantic similarities between motions.

\subsection{Ablation study}
\label{subsec:ablation}

To assess the effect of key design choices in BAL, we compare variants that remove or replace components of the proposed model in Tab.~\ref{tab:ablation}.
Pretraining and captioning follow the settings in the previous sections, while the forecasting and interpolation heads are trained for 57,200 iterations to reduce computational cost.
All variants use the same architecture for the representation extractor as BAL.
The pose-space variant additionally has a Perceiver IO-style decoder \cite{PerceiverIO}, and the JEPA~\cite{JEPA} variants use predictors with the same architecture as that used in BAL, with prediction targets obtained from an EMA-updated copy of the representation extractor.

\paragraph{Autoregression in latent space.}
BAL uses the natural ordering of daily human behavior as self-supervision by autoregressively predicting semantically abstract latent tokens.
Pose-space prediction instead forces the model to predict detailed future poses, even though the same movement semantics can yield different poses depending on context and individual style.
This ambiguity can make the objective less suitable for learning discriminative semantic representations.
Indeed, this variant achieves worse performance than BAL across all tasks.

Bidirectional masked prediction methods such as JEPA face a trade-off in mask length.
With short masks, past and future contexts complement each other in predicting the target, so neither direction alone needs to retain sufficient information for the prediction.
With long masks, positions far from the observed context admit multiple plausible actions as targets.
Autoregressive prediction, by contrast, requires a single direction to retain sufficient information to predict the next movement, which is tightly constrained by the observed history.
Its prediction errors therefore reflect a lack of information in the representations needed for prediction, rather than uncertainty among plausible actions, providing a more informative supervisory signal.
Consistent with this interpretation, BAL outperforms the JEPA variants across all tasks.

\paragraph{Decomposition into local and global components.}
Following SimSiam~\cite{SimSiam}, we analyze representation collapse by monitoring the training loss and feature standard deviation during pretraining (Fig.~\ref{fig:loss_std}).
For this analysis, we use only 10\% of the pretraining steps used in the full experiments to reduce computational cost.
The standard deviation is computed over L2-normalized representations and averaged over feature dimensions; it is zero under collapse to a constant and close to $1/\sqrt{d}$ for approximately zero-mean isotropic features in $d$ dimensions.
Without our decomposition, or when replacing it with SIGReg~\cite{SIGReg} as used in LeWM~\cite{LeWM}, the standard deviation remains low, indicating limited representation diversity and a tendency toward collapse.
Correspondingly, the training loss becomes very small in these variants, since such representations make the prediction objective easier to minimize.
Our decomposition maintains substantially higher diversity throughout pretraining, mitigating this tendency and leading to consistently better downstream performance. For SIGReg, we use the same number of random projections and regularization weight as Maes \etal~\cite{LeWM}.

Minimizing the loss only on the local components, which capture context-independent movement, degrades performance as it provides no supervision for sequence-level movement semantics.
Dynamically balancing the global loss weight prevents optimization from favoring either component and improves performance.
Directly predicting the global and local components also outperforms the A4Mer-style variant, which obtains these components by decomposing the predicted latent tokens.

\paragraph{Bidirectional autoregression.}
BAL uses both past-to-future and future-to-past autoregressive training to fully exploit ordering constraints while supporting bidirectional inference.
Bidirectional training improves performance on most metrics (Bi-AR/Causal vs. AR/Causal).
Removing the causal mask at inference time further improves performance on captioning and interpolation (Full, \ie, Bi-AR/Non-Causal vs. Bi-AR/Causal); for forecasting, we keep the causal mask in the extractor to prevent future information from leaking into the forecasting head, so the two settings coincide.
Mask-free inference even outperforms a two-pass alternative that concatenates representations extracted with causal and reverse causal masks (Full vs. Bi-AR/Bi-Causal). This result indicates that inferring the semantics of each motion chunk jointly from past and future context is more effective than processing the two contexts independently and combining them only after extraction.
The benefit from removing the causal masks does not emerge from past-to-future training alone (AR/Non-Causal vs. AR/Causal).

\begin{table}
    \centering
    \scriptsize
    \caption{
        Ablation studies of key design choices. ``A4Mer'': A4Mer-style local--global decomposition. ``(Bi-)AR'': (bidirectional) autoregressive training. ``Causal''/``Non-Causal''/``Bi-Causal'': inference with a causal mask, without a mask, or with concatenated causal and reverse causal outputs. ``Full'' corresponds to Bi-AR/Non-Causal.
        For forecasting, all variants retain the causal mask to prevent target information leakage; thus, the results marked with the same symbol ($\dag$ or $\ddag$) are identical. Motion captioning is evaluated on Nymeria.
        Parenthesized results report mean (standard deviation) over 10 runs; others are from a single run.
    }
    \label{tab:ablation}
    \vspace{5pt}
    \renewcommand{\arraystretch}{1.2}
    \setlength{\tabcolsep}{1.2pt}
    \begin{tabular}[t]{@{}l|l|cccc|ccc|ccc@{}}
        \toprule
        \multirow{2}{*}{\textbf{Key}}
        & \multirow{2}{*}{\textbf{Substitute}}
        & \multicolumn{4}{c|}{\textbf{Captioning} $\uparrow$}
        & \multicolumn{3}{c|}{\textbf{Forecasting} (mm) $\downarrow$}
        & \multicolumn{3}{c}{\textbf{Interpolation} (mm) $\downarrow$} \\
        &
        & BLEU-1 & R-L & CIDEr & BERT
        & AMD & Nymeria & E3D
        & AMD & Nymeria & E3D \\
        \midrule
        \multirow{3}{*}{Latent AR}
        & Pose-space Bi-AR                   & 17.6 & 18.9 & 12.6 & 7.3 & 118.6 & 96.6 & 64.7 & 125.7 & 87.4 & 60.1 \\
        & JEPA, mask 15\%                    & 34.5 & 39.0 & 46.2 & 25.2 & 109.3 & 107.9 & 91.6 & 145.8 & 112.4 & 83.4 \\
        & JEPA, mask 50\%                    & 32.0 & 38.4 & 41.7 & 24.9 & 102.7 & 99.0 & 86.7 & 133.5 & 96.8 & 82.1 \\
        \midrule
        \multirow{5}{*}{\shortstack[l]{Decompose \\ targets}}
        & w/o decomp.                        & 33.1 & 37.9 & 37.3 & 22.2 & 95.5 & 91.7 & 75.0 & 121.6 & 87.3 & 64.1 \\
        & SIGReg \cite{SIGReg,LeWM}          & 22.7 & 33.0 & 13.0 & 17.9 & 297.9 & 267.7 & 148.9 & 155.3 & 117.2 & 82.5 \\
        & w/o global                         & 27.4 & 33.4 & 12.4 & 16.6 & 172.9 & 131.9 & 104.3 & 149.0 & 129.0 & 84.3 \\
        & w/o balancing $\lambda$            & 24.7 & 28.8 & 49.6 & 4.2 & 170.0 & 180.2 & 105.5 & 157.1 & 225.7 & 80.9 \\
        & A4Mer \cite{A4Mer}                 & 21.7 & 35.3 & 24.3 & 18.6 & 107.2 & 92.0 & 69.8 & 126.5 & 89.0 & 66.2 \\
        \midrule
        \multirow{4}{*}{\shortstack[l]{Bi-AR /\\ Non-Causal}}
        & AR/Causal
        & 34.3\,(1.5)
        & 39.9\,(1.6)
        & 63.7\,(1.4)
        & 28.9\,(0.8)
        & \sbest 89.1\,(1.8)$^{\dag}$
        & \sbest 82.9\,(1.8)$^{\dag}$
        & \sbest 63.6\,(0.2)$^{\dag}$
        & 121.7\,(1.7)
        & 88.3\,(8.4)
        & 59.6\,(0.2) \\

        & AR/Non-Causal
        & 33.4\,(2.0)
        & 38.6\,(2.0)
        & 60.3\,(1.2)
        & 27.5\,(1.3)
        & \sbest 89.1\,(1.8)$^{\dag}$
        & \sbest 82.9\,(1.8)$^{\dag}$
        & \sbest 63.6\,(0.2)$^{\dag}$
        & \best 119.1\,(1.9)
        & 88.0\,(4.1)
        & 59.8\,(0.2) \\

        & Bi-AR/Causal
        & \sbest 35.1\,(0.9)
        & \sbest 40.8\,(0.6)
        & \sbest 65.2\,(1.5)
        & \sbest 29.6\,(0.8)
        & \best 88.3\,(1.3)$^{\ddag}$
        & \best 81.3\,(1.4)$^{\ddag}$
        & \best 63.2\,(0.1)$^{\ddag}$
        & 122.7\,(2.8)
        & 82.0\,(1.6)
        & \sbest 59.4\,(0.1) \\

        & Bi-AR/Bi-Causal
        & 34.2\,(1.2)
        & 39.9\,(0.7)
        & 62.6\,(1.9)
        & 29.0\,(0.7)
        & \best 88.3\,(1.3)$^{\ddag}$
        & \best 81.3\,(1.4)$^{\ddag}$
        & \best 63.2\,(0.1)$^{\ddag}$
        & 124.0\,(2.4)
        & \sbest 81.6\,(2.2)
        & 59.6\,(0.1) \\

        \midrule
        Full
        & ---
        & \best 36.5\,(1.2)
        & \best 41.2\,(0.4)
        & \best 68.3\,(1.3)
        & \best 30.8\,(0.8)
        & \best 88.3\,(1.3)$^{\ddag}$
        & \best 81.3\,(1.4)$^{\ddag}$
        & \best 63.2\,(0.1)$^{\ddag}$
        & \sbest 121.5\,(2.2)
        & \best 79.0\,(1.2)
        & \best 59.3\,(0.1) \\
        \bottomrule
    \end{tabular}
\end{table}

\section{Conclusion}

We proposed BAL, a self-supervised framework for learning movement representations that capture movement semantics from 3D pose sequences.
By autoregressively predicting in latent space, BAL turns the natural ordering structure inherent in everyday human behavior into self-supervision.
Bidirectional autoregressive prediction from past to future and from future to past enables BAL to fully exploit this structure and extract representations while taking both past and future context into account.
We further analyzed how decomposing prediction targets into global and local components helps prevent representation collapse, and empirically validated its effectiveness.
Experiments across diverse downstream tasks show that BAL's representations achieve competitive or superior downstream performance compared with existing self-supervised methods.
We hope that BAL will serve as a foundation for human movement representation learning and inspire research in broader domains beyond human movement.

\begin{ack}
This work was in part supported by
JSPS KAKENHI 
21H04893 and 26H02534,
and JST JPMJAP2305.
The authors declare no competing interests.
\end{ack}

{\small
\bibliographystyle{plainnat}
\bibliography{main}
}

\appendix

\section{Algorithm of BAL}
Alg.~\ref{alg:supp:bal} presents PyTorch-style pseudocode for BAL pretraining.

\begin{algorithm}[!t]
\caption{PyTorch-style pseudocode for BAL pretraining}
\label{alg:supp:bal}
\definecolor{codeblue}{rgb}{0.25,0.5,0.5}
\definecolor{codekw}{rgb}{0.85, 0.18, 0.50}
\lstset{
  backgroundcolor=\color{white},
  basicstyle=\fontsize{7.5pt}{7.5pt}\ttfamily\selectfont,
  columns=fullflexible,
  breaklines=true,
  captionpos=b,
  commentstyle=\fontsize{7.5pt}{7.5pt}\color{codeblue},
  keywordstyle=\fontsize{7.5pt}{7.5pt}\color{codekw},
  escapeinside={@@}{@@},
  keepspaces=true,
  numbers=none,
}

\begin{lstlisting}[language=python]
# f_E: Encoder
# f_L: LatentFormer
# g_f, g_b: forward and backward predictors
# D: distance function, including normalization
# B: batch size; T: number of frames; M: number of chunks
# d_in: input pose dimension; d: latent dimension

for X in loader: # [B, T, d_in]
    optimizer.zero_grad()

    h = f_E(X) # [B, M, d]

    z_f = f_L(h, mask="causal") # [B, M, d]
    z_b = f_L(h, mask="reverse_causal") # [B, M, d]

    z_l_f, z_g_f = decompose(z_f) # [B, M, d], [B, d]
    z_l_b, z_g_b = decompose(z_b) # [B, M, d], [B, d]

    # global components are predicted from a learnable CLS token
    pred_l_f, pred_g_f = g_f(z_f) # [B, M, d], [B, d]
    pred_l_b, pred_g_b = g_b(z_b) # [B, M, d], [B, d]

    # local loss
    loss_l_f = D(pred_l_f[:, 0:M-1, :], z_l_f.detach()[:, 1:M, :]) # [B, M-1]
    loss_l_b = D(pred_l_b[:, 1:M, :], z_l_b.detach()[:, 0:M-1, :]) # [B, M-1]

    # global loss
    loss_g_f = D(pred_g_f, z_g_f.detach()).unsqueeze(-1) # [B, 1]
    loss_g_b = D(pred_g_b, z_g_b.detach()).unsqueeze(-1) # [B, 1]

    lambda_f = (loss_l_f / loss_g_f).detach() # [B, M-1]
    lambda_b = (loss_l_b / loss_g_b).detach() # [B, M-1]

    loss_f = (loss_l_f + lambda_f * loss_g_f).mean()
    loss_b = (loss_l_b + lambda_b * loss_g_b).mean()
    loss = loss_f + loss_b

    loss.backward() # back-propagate
    optimizer.step() # update parameters

def decompose(z):
    z_global = z.mean(dim=1) # [B, d]
    z_local = z - z_global[:, None, :] # [B, M, d]
    return z_local, z_global

def D(pred, target, is_l2_norm=False):
    if is_l2_norm:
        pred = normalize(pred, dim=-1)
        target = normalize(target, dim=-1)
        loss = 1 - (pred * target).sum(dim=-1)

    else:
        # pred is left unnormalized
        target = layer_norm(target, dim=-1)
        loss = smooth_l1(pred, target, reduction="none").mean(dim=-1)

    return loss
\end{lstlisting}
\end{algorithm}

\section{Diversity of normalized local prediction targets}
\label{subsec:supp:decompose}
In Sec.~\ref{subsec:decompose}, we established an angular separation bound for L2-normalized local components when all local components are nonzero. Here, we extend this result to zero local components and derive a corresponding bound for layer normalization~\cite{LayerNorm}, including the effect of numerical stabilization.
For $M\ge 2$ latent tokens in $\mathbb{R}^d$ with $d\ge 2$, the local components satisfy
\begin{equation}
    z_m^\lcl=z_m-\frac{1}{M}\sum_{n=1}^M z_n\,,\qquad
    \sum_{m=1}^M z_m^\lcl=0\,.
    \label{eq:supp:local-zero-sum}
\end{equation}

\subsection{L2 normalization with zero local components}
Suppose the latent tokens are not all identical, and let $K$ denote the number of nonzero local components. The zero-sum property in Eq.~(\ref{eq:supp:local-zero-sum}) implies $K\ge 2$, since a single nonzero component cannot sum to zero.
Without loss of generality, let these be the first $K$ components, and define
\begin{equation}
    a_m=\|z_m^\lcl\|>0\,,\qquad
    u_m=\frac{z_m^\lcl}{a_m}\,.
\end{equation}
Since $z_m^\lcl=a_m u_m$, Eq.~(\ref{eq:supp:local-zero-sum}) gives
\begin{equation}
    \sum_{m=1}^K a_m u_m
    =\sum_{m=1}^M z_m^\lcl=0\,.
    \label{eq:supp:l2-zero-sum}
\end{equation}
Let $\mu=\min_{1\le m<n\le K}u_m^\top u_n$. Taking the squared norm of Eq.~(\ref{eq:supp:l2-zero-sum}) yields
\begin{align}
    0
    &=\left\|\sum_{m=1}^K a_m u_m\right\|^2 \notag\\
    &=\sum_{m=1}^K a_m^2
      +2\sum_{m<n}a_m a_n\,u_m^\top u_n \notag\\
    &\ge \sum_{m=1}^K a_m^2
      +2\mu\sum_{m<n}a_m a_n\,.
    \label{eq:supp:l2-lower}
\end{align}
Cauchy--Schwarz also gives
\begin{align}
    2\sum_{m<n}a_m a_n
    &=\left(\sum_{m=1}^K a_m\right)^2
      -\sum_{m=1}^K a_m^2 \notag\\
    &\le (K-1)\sum_{m=1}^K a_m^2\,.
\end{align}
Combining these inequalities, we obtain
\begin{equation}
    \mu\le -\frac{\sum_{m=1}^K a_m^2}{2\sum_{m<n}a_m a_n}
        \le -\frac{1}{K-1}\,.
    \label{eq:supp:l2-bound}
\end{equation}
As $\arccos$ is decreasing,
\begin{equation}
    \max_{m<n}\angle(u_m,u_n)
    \ge \arccos\!\left(-\frac{1}{K-1}\right)
    >\frac{\pi}{2}\,.
    \label{eq:supp:l2-angle}
\end{equation}
Thus, the nonzero L2-normalized local components necessarily contain a pair separated by more than $90^\circ$. Zero local components are excluded because their L2 normalization is undefined. For $K=M$, this recovers the result in Sec.~\ref{subsec:decompose}.

\subsection{Layer normalization}
We consider layer normalization without affine parameters, as used for the prediction targets to which stop-gradient is applied.
Let $P$ denote the centering matrix that subtracts the mean across feature dimensions:
\begin{equation}
    \mathbf{1}=(1,\ldots,1)^\top\in\mathbb{R}^d\,,\qquad
    P=I_d-\frac{1}{d}\mathbf{1}\mathbf{1}^\top\,.
\end{equation}
We denote layer normalization with a numerical-stability constant $\epsilon\ge 0$ by $\operatorname{LN}_\epsilon$, defined as
\begin{equation}
    \operatorname{LN}_\epsilon(x)
    =\frac{Px}{\sqrt{\|Px\|^2/d+\epsilon}}\,.
    \label{eq:supp:ln-def}
\end{equation}
For $\epsilon=0$, this operation is defined only when $Px\ne 0$.
By linearity of $P$, Eq.~(\ref{eq:supp:local-zero-sum}) implies
\begin{equation}
    \sum_{m=1}^M Pz_m^\lcl
    =P\left(\sum_{m=1}^M z_m^\lcl\right)=0\,.
    \label{eq:supp:ln-zero-sum}
\end{equation}
Let $K$ now denote the number of nonzero feature-centered local components. If $K=0$, all outputs are zero for $\epsilon>0$ and undefined for $\epsilon=0$. This can occur even when the latent tokens are not identical, since $P$ removes components that are constant across feature dimensions. If $K>0$, the zero-sum property in Eq.~(\ref{eq:supp:ln-zero-sum}) implies $K\ge 2$.

Without loss of generality, let the first $K$ feature-centered components be nonzero, and define
\begin{equation}
    b_m=\|Pz_m^\lcl\|>0\,,\qquad
    w_m=\frac{Pz_m^\lcl}{b_m}\,,\qquad
    v_m=\frac{b_m^2}{d}\,.
\end{equation}
Here, $v_m$ is the variance of $z_m^\lcl$ across feature dimensions. Since
\begin{equation}
    \sum_{m=1}^K b_m w_m
    =\sum_{m=1}^M Pz_m^\lcl=0\,,
\end{equation}
the argument in the preceding subsection applies to these feature-centered components, yielding
\begin{equation}
    \min_{m<n}w_m^\top w_n
    \le -\frac{1}{K-1}\,.
    \label{eq:supp:ln-angular}
\end{equation}
From Eq.~(\ref{eq:supp:ln-def}), the normalized outputs can be written as
\begin{equation}
    \operatorname{LN}_\epsilon(z_m^\lcl)=r_m w_m\,,\qquad
    r_m=\sqrt{\frac{dv_m}{v_m+\epsilon}}\,.
    \label{eq:supp:ln-radius}
\end{equation}
Let $\nu=\min_{1\le m\le K}v_m>0$. Then
\begin{equation}
    r_m\ge r_{\min}
    =\sqrt{\frac{d\nu}{\nu+\epsilon}}\,.
\end{equation}
For a pair $m<n$ satisfying Eq.~(\ref{eq:supp:ln-angular}), we obtain
\begin{align}
    \|r_m w_m-r_n w_n\|^2
    &=r_m^2+r_n^2-2r_mr_n w_m^\top w_n\notag\\
    &\ge r_m^2+r_n^2+\frac{2r_mr_n}{K-1}\notag\\
    &\ge \frac{2dK}{K-1}\,\frac{\nu}{\nu+\epsilon}\,.
    \label{eq:supp:ln-distance-derivation}
\end{align}
Consequently, the maximum pairwise distance among the normalized nonzero feature-centered local components satisfies
\begin{equation}
    \begin{aligned}
    D_\epsilon
    &:=\max_{1\le m<n\le K}\|r_mw_m-r_nw_n\|\\
    &\ge\sqrt{\frac{2dK}{K-1}\,\frac{\nu}{\nu+\epsilon}}\,.
    \end{aligned}
    \label{eq:supp:ln-distance}
\end{equation}

To interpret this bound, consider first $\epsilon=0$, for which every defined layer-normalized local component has norm $\sqrt{d}$ by Eq.~(\ref{eq:supp:ln-radius}). These components lie in the zero-mean feature subspace, where the maximum possible distance between two vectors of this norm is $2\sqrt{d}$, attained by a pair of opposite vectors. Dividing Eq.~(\ref{eq:supp:ln-distance}) by this maximum gives
\begin{equation}
    \frac{D_0}{2\sqrt{d}}
    \ge \sqrt{\frac{K}{2(K-1)}}
    \ge \frac{1}{\sqrt{2}}\approx
    0.707\,.
    \label{eq:supp:ln-relative}
\end{equation}
Thus, when $\epsilon=0$, at least one pair of layer-normalized local targets is separated by at least approximately $70.7\%$ of the maximum possible distance. This lower bound is independent of both the number of tokens and the magnitudes of their nonzero feature-centered local components. In particular, the separation persists as the latent tokens approach one another, provided these components remain nonzero.

For $\epsilon>0$, Eq.~(\ref{eq:supp:ln-distance}) gives the corresponding bound with the factor $\sqrt{\nu/(\nu+\epsilon)}$. The separation therefore remains close to the $\epsilon=0$ bound when the nonzero feature-wise variances are large relative to the stability constant. Together, these results show that local--global decomposition maintains nontrivial diversity among normalized local prediction targets even when the underlying representations are similar. Near-constant predictions cannot simultaneously match such separated targets, helping to counteract the self-reinforcing collapse mechanism described in Sec.~\ref{subsec:decompose}.

\section{Implementation details}
\label{sec:supp:impldetail}

\subsection{Text preprocessing for motion captioning}
\label{subsec:supp:textPreprocess}
Nymeria \cite{Nymeria} provides text annotations at three levels of granularity.
We use the coarsest annotation level, ``activity summarization,'' which describes high-level activities rather than fine-grained motion details.
Since these annotations refer to the actor as ``C,'' we replace ``C'' with ``the person'' to make the text more natural and easier for the LLM to process.

Embody 3D (E3D) \cite{Embody3D} also provides text annotations at multiple levels of granularity.
As with Nymeria, we use the coarsest annotation level, ``describe person action.''

\subsection{Pose representation}
\label{subsec:supp:poseRepr}
Following HumanML3D \cite{HumanML3D}, we use an input pose representation in which body shape is normalized and each pose is expressed in a gravity-aligned local coordinate frame.
Each pose is represented by
$\left(\bm{c},\, \dot{r}^{a},\, \dot{\bm{r}}^{v},\, \bm{j},\, \bm{j}^{v},\, \bm{j}^{r}\right)$,
where $\bm{c}\in\{0,1\}^{4}$ indicates foot-ground contacts for the left and right heels and toes,
$\dot{r}^{a}\in\real$ denotes the pelvis angular velocity around the gravity axis,
and $\dot{\bm{r}}^{v}\in\real^{2}$ denotes the pelvis velocity on the ground plane.
The terms $\bm{j}\in\real^{J\times 3}$, $\bm{j}^{v}\in\real^{J\times 3}$, and $\bm{j}^{r}\in\real^{J\times 6}$ correspond to the joint positions, joint velocities, and joint rotations, respectively, with $J$ denoting the number of joints.
We use the continuous 6D rotation representation \cite{OnTheContinuityOfRotationRepresentations} for $\bm{j}^{r}$.
For baseline methods, we use the pose representation of each original paper so that each baseline is evaluated with the representation it was designed for.

\subsection{Network architecture}
\label{subsec:supp:arch}
We summarize the network architectures of BAL, the decoder, the forecasting head, the interpolation head, and the MLP for captioning in Tab.~\ref{tab:supp:architecture}. The task-specific modules share the same architecture across all methods. Since the representation dimensionality of the baselines varies and differs from that of BAL, we use a learnable linear layer to project each baseline representation to 256 dimensions before the decoder, forecasting head, and interpolation head. For captioning, we instead adjust the input dimension of the two-layer MLP to match each representation, while keeping its hidden and output dimensions unchanged.

\begin{table}[t]
\centering
\caption{Network architectures. All Transformer blocks use layer normalization (LN) \cite{LayerNorm}, width 256, 4 attention heads, MLP width 512, GELU activations \cite{GELU}, QK normalization \cite{QKNorm} with a learnable scale and no bias, RoPE \cite{RoPE} with $\theta=1000$, and LayerScale \cite{LayerScale}. For the head modules listed in the Module column, \#Layers denotes the number of linear layers.}
\label{tab:supp:architecture}
\vspace{5pt}
\footnotesize
\renewcommand{\arraystretch}{1.2}
\setlength{\tabcolsep}{1.8pt}
\begin{tabular}{@{}llccc@{}}
\toprule
Model & Module & Token Dim. & \#Layers & Component \\
\midrule

\multirow{4}{*}{BAL}
& Encoder & 256 & 8 & Transformer blocks \\
& LatentFormer & 256 & 8 & Transformer blocks \\
& Predictor & 256 & 3 & Transformer blocks \\
& Predictor head & 256 & 2 & LN + linear layers \\

\midrule

\multirow{2}{*}{Decoder}
& Decoder model & 256 & 4 & Transformer blocks \\
& Head & 256 & 1 & LN + linear layer \\

\midrule

\multirow{2}{*}{Forecasting head}
& Forecasting model & 256 & 4 & Transformer blocks \\
& Head & 256 & 1 & LN + linear layer \\

\midrule

\multirow{2}{*}{Interpolation head}
& Interpolation model & 256 & 4 & Transformer blocks \\
& Head & 256 & 1 & LN + linear layer \\

\midrule

\multirow{1}{*}{Captioning}
& MLP & $256 \rightarrow 2176 \rightarrow 4096$ & 2 & Linear + GELU + linear \\

\bottomrule
\end{tabular}
\end{table}

\paragraph{Normalization and distance function.}
For the distance function in Eqs.~(\ref{eq:fwd}), (\ref{eq:bwd}), (\ref{eq:local}), and (\ref{eq:global}), we use layer normalization and the smooth L1 distance, rather than L2 normalization and cosine dissimilarity, as this choice yields better downstream performance (Tab.~\ref{tab:supp:L2vsLayerNorm}).
With L2 normalization, both the prediction and the target are normalized.
With layer normalization, in contrast, we normalize only the target and leave the predictor outputs unnormalized for distance computation, allowing the final linear layer of the predictor to learn the affine correction needed to match the target distribution.

One possible explanation for the performance gap lies in how the two objectives constrain the predictor outputs.
With L2 normalization and cosine dissimilarity, the loss is invariant to positive rescaling of the predictor output, leaving its magnitude unconstrained.
In contrast, the smooth L1 distance directly penalizes coordinate-wise deviations of the unnormalized predictor output from the layer-normalized target.
For a fixed target, the objective is uniquely minimized when the predictor output matches the target, thus constraining its magnitude as well as its direction.

\begin{table*}[t!]
    \centering
    \caption{Comparison of normalization operations and distance functions across motion captioning, forecasting, and interpolation.
    Layer normalization with the smooth L1 distance yields better downstream performance than L2 normalization with cosine dissimilarity across all tasks.}
    \label{tab:supp:L2vsLayerNorm}
    \fontsize{8pt}{9pt}\selectfont
    \renewcommand{\arraystretch}{1.2}
    \setlength{\tabcolsep}{1.8pt}
    \begin{tabular}{@{}ll|cccccccc|ccc|ccc@{}}
    \toprule
    \multirow{3}{*}{\textbf{Normalization}} & \multirow{3}{*}{\textbf{Distance}}
    & \multicolumn{8}{c|}{\textbf{Captioning} $\uparrow$}
    & \multicolumn{3}{c|}{\multirow{2}{*}{\textbf{Forecasting} (mm) $\downarrow$}}
    & \multicolumn{3}{c}{\multirow{2}{*}{\textbf{Interpolation} (mm) $\downarrow$}} \\
    & & \multicolumn{2}{c}{BLEU-1}
      & \multicolumn{2}{c}{ROUGE-L}
      & \multicolumn{2}{c}{CIDEr}
      & \multicolumn{2}{c|}{BERTScore}
      & \multicolumn{3}{c|}{}
      & \multicolumn{3}{c}{} \\
    & & Nym & E3D
      & Nym & E3D
      & Nym & E3D
      & Nym & E3D
      & AMD & Nym & E3D
      & AMD & Nym & E3D \\
    \midrule
    L2 norm & Cosine & 28.9 & 58.3 & 39.4 & 44.7 & 50.8 & 17.1 & 27.2 & 33.9 & 96.7 & 84.6 & 68.8 & 121.8 & 81.7 & 63.1 \\
    Layer norm    & Smooth L1    & \best 37.3 & \best 58.9 & \best 41.2 & \best 45.3 & \best 63.1 & \best 18.6 & \best 29.6 & \best 35.0 & \best 90.5 & \best 82.1 & \best 63.8 & \best 117.5 & \best 77.0 & \best 59.2 \\
    \bottomrule
    \end{tabular}
\end{table*}

\subsection{Training settings}
\label{subsec:supp:trainSettings}
We summarize the training settings for BAL pretraining and each downstream task in Tab.~\ref{tab:supp:training}. In all downstream tasks, the pretrained representation extractors, including those of the baselines, are kept frozen, and only the task-specific modules are trained. These task-specific modules are trained on the same dataset and with the same settings for all methods, so that the downstream results reflect the quality of the learned representations rather than differences in training data or configurations.

\begin{table}[t]
\centering
\caption{Training settings. All modules listed in the table are optimized with AdamW \cite{AdamW} and gradient norm clipping at 1.0, using linear warmup followed by a cosine learning-rate schedule \cite{SGDR}. For BAL, the weight decay is increased from 0.04 to 0.4 following a cosine schedule. All modules are trained on a single NVIDIA A100 GPU.}
\label{tab:supp:training}
\vspace{5pt}
\footnotesize
\renewcommand{\arraystretch}{1.2}
\setlength{\tabcolsep}{1.8pt}
\begin{tabular}{@{}lcccccc@{}}
\toprule
\textbf{Task} & \textbf{Batch size} & \textbf{Learning rate} & \textbf{Weight decay} & \textbf{Iterations} & \textbf{Warmup iter.} & \textbf{Training time (h)} \\
\midrule
BAL
& 128
& $5 \times 10^{-4}$
& $0.04 \rightarrow 0.4$
& 429K
& 2000
& 39.5 \\

Decoder
& 128
& $1 \times 10^{-3}$
& 0.01
& 143K
& 2000
& 4.9 \\

Forecasting
& 128
& $1 \times 10^{-3}$
& 0.01
& 286K
& 2000
& 9.6 \\

Interpolation
& 128
& $1 \times 10^{-3}$
& 0.01
& 143K
& 2000
& 21.5 \\

Captioning
& 64
& $8 \times 10^{-5}$
& 0.0
& 2.6K
& 1000
& 2.3 \\

\bottomrule
\end{tabular}
\end{table}

\paragraph{Motion captioning.}
For motion captioning, we project the movement representations extracted by the representation extractor into the language embedding space of an LLM using a two-layer MLP.
We then construct the input to the LLM by concatenating the system-prompt embeddings, the projected movement embeddings, and the user-prompt embeddings in this order.
We use the following system prompt: ``You are an assistant that describes human motion concisely in natural English. Describe only what is observable from the given motion.''
For the user prompt, we use: ``Describe this human motion sequence in natural English as a short paragraph in chronological order.''

We train only the projection MLP with the standard autoregressive negative log-likelihood loss, keeping the representation extractor and the LLM frozen. The loss is computed only over the ground-truth caption tokens; the system prompt, user prompt, and projected movement embeddings are excluded.

\paragraph{Decoder.}
Since motion forecasting and interpolation are performed in the latent space, we introduce a decoder that maps latent tokens to poses so that the predictions can be evaluated in pose space.
The decoder is implemented as a simple Transformer Encoder \cite{Transformer}.
Its input is obtained by repeating each latent token over the frames of the chunk it represents, yielding a frame-level sequence of length $T$; the decoder then maps this sequence to the corresponding poses.
Following A4Mer \cite{A4Mer}, we train the decoder with a reconstruction objective $\loss_\mathrm{dec}$ that combines joint-position and velocity losses:
\begin{align}
    \loss_{\mathrm{dec}}
    &= \lambda_\mathrm{pos}\loss_{\mathrm{pos}}
    + \lambda_\mathrm{vel}\loss_{\mathrm{vel}}\,, \\
    \loss_{\mathrm{pos}}
    &= \frac{1}{J T}\sum_{t=1}^T \sum_{j=1}^J
    \left\| \hat{\bm{j}}_{j,t} - \bm{j}_{j,t} \right\|\,, \\
    \loss_{\mathrm{vel}}
    &= \frac{1}{J (T-1)}\sum_{t=2}^{T} \sum_{j=1}^J
    \left\| \hat{\Delta\bm{j}}_{j,t} - \Delta\bm{j}_{j,t} \right\|\,,
\end{align}
where $\bm{j}_{j,t}$ and $\hat{\bm{j}}_{j,t}$ denote the ground-truth and reconstructed positions of joint $j$ at frame $t$, respectively.
The velocities are defined as
$\Delta\bm{j}_{j,t} = \bm{j}_{j,t} - \bm{j}_{j,t-1}$
and
$\hat{\Delta\bm{j}}_{j,t} = \hat{\bm{j}}_{j,t} - \hat{\bm{j}}_{j,t-1}$.
We set $\lambda_\mathrm{pos}=1.5$ and $\lambda_\mathrm{vel}=1.0$.

\paragraph{Motion forecasting} is performed by predicting the next latent token $z^\fwd_{m+1}$ from the latent token sequence $z^\fwd_{\le m}$ produced by the pretrained representation extractor, and then decoding the predicted token into pose space.
Following A4Mer, we train the autoregressive forecasting head using a latent-token forecasting loss $\loss_\mathrm{token}$ and a joint-space reconstruction loss $\loss_\mathrm{dec}$ computed by decoding the predicted latent tokens with the frozen decoder:
\begin{align}
    \loss_{\mathrm{fore}} &= \loss_{\mathrm{token}} + \loss_{\mathrm{dec}}\,, \\
    \loss_{\mathrm{token}} &= \frac{1}{M-1}\sum_{m=1}^{M-1} \mathrm{SmoothL1}\left(\hat{z}^\fwd_{m+1} - z^\fwd_{m+1}\right)\,,
\end{align}
where $\hat{z}^\fwd_{m+1}$ denotes the latent token predicted by the head.
The reconstruction term helps prevent the forecasting head from optimizing latent dimensions that are not relevant to reconstructing plausible movement.

When A4Mer is used as the representation extractor, its latent tokens correspond to variable-length chunks. Since the lengths of future chunks are unavailable at inference time, we follow the original A4Mer protocol and train the forecasting head to estimate the next chunk length as well. Because the distribution of chunk lengths has a long right tail, we regress them in log space:
\begin{align}
    \loss_{\mathrm{len}} = \frac{1}{M-1}\sum_{m=1}^{M-1} \left| \log l_{m+1} - \mathrm{Softplus}\left(\hat{l}_{m+1}\right) \right|\,,
\end{align}
where $\hat{l}_{m+1}$ denotes the raw output of the head and $l_{m+1}$ denotes the chunk length determined by the pretrained A4Mer model rather than by manual annotation.
In this case, the final forecasting loss is
\begin{align}
    \loss_{\mathrm{fore}} = \loss_{\mathrm{token}} + \loss_{\mathrm{dec}} + \loss_{\mathrm{len}}\,.
\end{align}
During training, the predicted latent tokens are decoded with the ground-truth chunk lengths. At inference time, we recover the estimated chunk lengths by exponentiating the predicted log lengths and use them for decoding.

\paragraph{Motion interpolation} is performed by predicting the latent tokens corresponding to unobserved frames of the input sequence and decoding the completed latent token sequence into pose space.
We first extract latent tokens from the masked input sequence, in which the masked frames are excluded from attention through an attention mask. We then replace the latent tokens at the positions corresponding to the masked regions with learnable tokens and feed the resulting sequence to an interpolation head.

For methods with fixed-length chunks, the latent positions associated with the masked regions are determined directly from the temporal grid.
In contrast, A4Mer produces latent tokens from an adaptive segmentation, so its tokens correspond to variable-length chunks.
A strict interpolation setup for A4Mer would therefore require inferring the adaptive chunk partition within each masked region, including both the number of missing chunks and their lengths.
Since this introduces a nontrivial extension of A4Mer, we instead determine the chunk partition by applying the pretrained A4Mer extractor to the fully observed sequence.
Note that this gives A4Mer access to the chunk partition of the unobserved frames, which favors A4Mer in this comparison.

We train the interpolation head with a latent reconstruction loss $\loss_\mathrm{latent}$ between the predicted latent tokens $\hat{z}_m$ and the target latent tokens $z_m$ extracted from the fully observed sequence by the representation extractor without causal masking:
\begin{align}
    \loss_{\mathrm{latent}}
    =
    \frac{1}{M}\sum_{m=1}^{M}
    \mathrm{SmoothL1}\left(\hat{z}_{m} - z_{m}\right)\,.
\end{align}
Although the main objective is to recover the latent tokens of the masked regions, we compute the loss over all latent positions to account for the distributional discrepancy between latent tokens extracted from masked and fully observed inputs.
As in motion forecasting, we further add the joint-space reconstruction loss $\loss_{\mathrm{dec}}$ by decoding the completed latent token sequence with the frozen decoder.
The final interpolation loss is
\begin{align}
    \loss_{\mathrm{interp}}
    =
    \loss_{\mathrm{latent}}
    +
    \loss_{\mathrm{dec}}\,.
\end{align}

\section{Additional experiments}

\subsection{XLNet}
To assess whether bidirectional autoregressive prediction along the natural temporal order provides more effective self-supervision than prediction over random factorization orders, as in XLNet \cite{XLNet}, we compare BAL with a model trained using an XLNet-style objective.

To isolate the effect of the prediction objective as much as possible, we retain BAL's Encoder for producing chunk embeddings and replace the LatentFormer and predictors with XLNet-style two-stream attention comprising a content stream and a query stream.
The two streams share a single set of parameters, and the resulting network has the same depth, width, and number of attention heads as the LatentFormer. All other training settings are also identical to those of BAL.
In this network, the content stream encodes both the context and each token itself as in standard self-attention, whereas the query stream encodes the target position and aggregates information from the content stream of the tokens preceding the target in the factorization order, without accessing the content of the target token.
The final query-stream output is passed to the same predictor head as in BAL (see Tab.~\ref{tab:supp:architecture}) to predict the final content-stream representation of the target token. The target is detached from the computational graph, and the loss is computed in latent space as in BAL.
Following XLNet, approximately one sixth of the latent tokens in each sequence are selected as prediction targets in the form of contiguous spans of at most five tokens.
At inference time, we discard the query stream and extract representations from the content stream without permutation masking.

As shown in Tab.~\ref{tab:supp:xlnet}, BAL outperforms the XLNet-style model, particularly on motion captioning---the task that most directly reflects whether the learned representations capture movement semantics. This suggests that random factorization orders do not sufficiently exploit the natural ordering structure of everyday behavior as self-supervision, and that autoregressive prediction along the natural temporal order is important for learning meaningful movement representations.

\begin{table*}[t!]
    \centering
    \footnotesize
    \caption{Comparison with an XLNet-style variant. BAL outperforms it on most metrics.}
    \label{tab:supp:xlnet}
    \renewcommand{\arraystretch}{1.2}
    \setlength{\tabcolsep}{3pt}
    \begin{tabular}{@{}l|cccccccc|ccc|ccc@{}}
    \toprule
    \multirow{3}{*}{\textbf{Method}} & \multicolumn{8}{c|}{\textbf{Captioning} $\uparrow$} & \multicolumn{3}{c|}{\multirow{2}{*}{\textbf{Forecasting} (mm) $\downarrow$}} & \multicolumn{3}{c}{\multirow{2}{*}{\textbf{Interpolation} (mm) $\downarrow$}} \\
    & \multicolumn{2}{c}{BLEU-1} & \multicolumn{2}{c}{ROUGE-L} & \multicolumn{2}{c}{CIDEr} & \multicolumn{2}{c|}{BERTScore} & \multicolumn{3}{c|}{} & \multicolumn{3}{c}{} \\
    & Nym & E3D & Nym & E3D & Nym & E3D & Nym & E3D & AMD & Nym & E3D & AMD & Nym & E3D \\
    \midrule
    XLNet         &       35.2  &       57.9 &       38.6  &      43.5 &       42.9 &       16.3 &       26.7 &       32.3 & \best 85.8 &       87.1 &       65.6 & \best 116.5 &       79.3  & \best 58.1 \\
    \textbf{BAL}  & \best 37.3  & \best 58.9 & \best 41.2 & \best 45.3 & \best 63.1 & \best 18.6 & \best 29.6 & \best 35.0 &       90.5 & \best 82.1 & \best 63.8 &       117.5 & \best 77.0 &         59.2  \\
    \bottomrule
    \end{tabular}
\end{table*}

\subsection{Effect of global translation on motion forecasting and interpolation}
\label{subsec:supp:largeError}
We hypothesize that the relatively large errors of the baseline methods in motion forecasting and interpolation are partly attributable to their learned representations preserving global translation information less effectively than those learned by BAL.

To test this hypothesis, we conduct an oracle global-translation alignment experiment.
For each forecasted or interpolated sequence, we compute the pelvis offset between the first predicted frame and its ground-truth counterpart and translate the entire predicted sequence by this offset.
For each method and evaluation setting, we report the lower of the original and aligned errors, ensuring that the alignment does not penalize methods for which it is unhelpful.

Tab.~\ref{tab:supp:largeError} reports the results of this oracle alignment experiment.
Compared with the original errors in Tab.~\ref{tab:results}, the alignment substantially reduces the performance gap between BAL and the baselines, providing evidence that global translation errors account for part of this gap.

Importantly, however, this alignment does not eliminate BAL's advantage.
Even under this oracle alignment, which particularly benefits methods with large global translation errors, BAL still outperforms the baselines in most evaluation settings, and the pairwise ordering between BAL and each baseline remains unchanged from Tab.~\ref{tab:results}.
Thus, although global translation errors contribute to the baselines' larger errors, they do not fully explain BAL's advantage.

Note that because this alignment relies on ground-truth poses, which are unavailable at inference time, the original results in Tab.~\ref{tab:results} remain the appropriate measure of practical forecasting and interpolation performance.

\begin{table}[t!]
    \centering
    \footnotesize
    \caption{Effect of oracle global-translation alignment on motion forecasting and interpolation. BAL remains the best in most settings after alignment.}
    \vspace{5pt}
    \label{tab:supp:largeError}
    \renewcommand{\arraystretch}{1.2}
    \setlength{\tabcolsep}{3.0pt}
    \begin{tabular}{@{}l|ccc|ccc@{}}
    \toprule
    \multirow{2}{*}{\textbf{Method}}
    & \multicolumn{3}{c|}{\textbf{Forecasting} (mm) $\downarrow$}
    & \multicolumn{3}{c}{\textbf{Interpolation} (mm) $\downarrow$} \\
    & AMD & Nym & E3D
    & AMD & Nym & E3D \\
    \midrule
    A4Mer
    & 633.9
    & 285.1
    & 375.5
    & 163.7
    & 120.8
    & 86.3 \\

    PUMPS
    & 163.2
    & 130.3
    & 102.2
    & 182.6
    & 154.3
    & 87.0 \\

    USDRL
    & \sbest 133.3
    & 115.9
    & 87.9
    & 117.7
    & 106.7
    & \sbest 59.2 \\

    BehaveMAE
    & 170.2
    & 169.1
    & 126.1
    & 135.6
    & 103.4
    & 77.8 \\

    MotionBERT
    & 282.4
    & 244.1
    & 162.0
    & 135.2
    & 129.8
    & 69.2 \\

    MacDiff
    & 152.7
    & \sbest 77.9
    & \sbest 71.5
    & \best 106.9
    & \sbest 100.4
    & \best 55.4 \\
    \midrule
    \textbf{BAL}
    & \best 87.8
    & \best 77.7
    & \best 63.7
    & \sbest 117.5
    & \best 77.0
    & \sbest 59.2 \\
    \bottomrule
    \end{tabular}
\end{table}

\subsection{Effect of chunk length}
\label{subsec:supp:chunk}
In BAL, we set the chunk length to 4 seconds based on the definition of atomic actions in Nymeria \cite{Nymeria}.
The chunk length controls the temporal scale represented by each latent token and is the only hyperparameter in BAL involved in the design of the supervisory signal.
Unlike augmentation strengths or masking scales in prior methods, however, it has a direct semantic interpretation: it specifies the temporal unit over which movement is abstracted, and can therefore be chosen intuitively based on the expected duration of meaningful actions.

To examine how this temporal granularity affects the learned representations, we pretrain BAL with different chunk lengths and evaluate the resulting representations on all downstream tasks.
As shown in Tab.~\ref{tab:supp:chunk}, performance is consistently strong across the tested range, with every variant either outperforming existing self-supervised methods or remaining competitive with them. Thus, BAL achieves strong downstream performance without requiring careful tuning of the chunk length for each task.

The results also reveal a modest dependence on the task.
Shorter chunks generally perform better for motion forecasting and interpolation, whereas longer chunks tend to benefit motion captioning.
This tendency is consistent with the nature of the downstream tasks.
Because forecasting and interpolation are evaluated directly in pose space, they benefit from fine-grained pose information, which shorter chunks are more likely to retain.
In contrast, captioning requires higher-level semantic understanding of movements, for which longer temporal units can provide a more suitable abstraction.
When further gains are desired, the chunk length can therefore be chosen according to an intuitive notion of the temporal granularity relevant to the target task.

\begin{table}[t!]
    \centering
    \caption{Effect of the chunk length on motion captioning, forecasting, and interpolation. ``Best prior result'' denotes, for each metric, the best result among prior self-supervised methods. Across the tested chunk lengths, BAL outperforms or remains competitive with the best prior result. We use a 4-second chunk length in the main text.}
    \label{tab:supp:chunk}
    \vspace{5pt}
    \fontsize{8.3pt}{10pt}\selectfont
    \renewcommand{\arraystretch}{1.2}
    \setlength{\tabcolsep}{1.8pt}
    \begin{tabular}{@{}l|l|cccccccc|ccc|ccc@{}}
    \toprule
    \multirow{3}{*}{\textbf{Method}} & \multirow{3}{*}{\shortstack{\textbf{Chunk}\\\textbf{length}}}
    & \multicolumn{8}{c|}{\textbf{Captioning} $\uparrow$}
    & \multicolumn{3}{c|}{\multirow{2}{*}{\textbf{Forecasting} (mm) $\downarrow$}}
    & \multicolumn{3}{c}{\multirow{2}{*}{\textbf{Interpolation} (mm) $\downarrow$}} \\
    & & \multicolumn{2}{c}{BLEU-1}
      & \multicolumn{2}{c}{ROUGE-L}
      & \multicolumn{2}{c}{CIDEr}
      & \multicolumn{2}{c|}{BERTScore}
      & \multicolumn{3}{c|}{}
      & \multicolumn{3}{c}{} \\
    & & Nym & E3D
      & Nym & E3D
      & Nym & E3D
      & Nym & E3D
      & AMD & Nym & E3D
      & AMD & Nym & E3D \\
    \midrule
    BAL                & 2 sec & \sbest 37.4 & 58.7        & 41.0        & 45.0        & 62.6        & 16.9        & 28.0        & 34.2        & \best 89.1   & \best 72.2  & \best 61.4  & \best 110.6  & \best 70.3  & \best 55.0  \\
    BAL                & 3 sec & 33.4        & \sbest 58.9 & 40.2        & 45.1        & 63.6        & 17.7        & 28.7        & 34.6        & 95.1         & 82.2        & \sbest 63.4 & \sbest 111.3 & \sbest 75.2 & \sbest 56.3 \\
    BAL                & 4 sec & 37.3        & \sbest 58.9 & 41.2        & \best 45.3  & 63.1        & \best 18.6  & \sbest 29.6 & \sbest 35.0 & \sbest 90.5  & \sbest 82.1 & 63.8        & 117.5        & 77.0        & 59.2        \\
    BAL                & 5 sec & \best 41.4  & \best 59.1  & \best 41.7  & \sbest 45.2 & \best 71.8  & 18.3        & \best 31.6  & 34.5        & 115.3        & 87.5        & 66.4        & 118.3        & 80.7        & 59.6        \\
    Best prior result  &   -   & 36.7        & 58.8        & \sbest 41.4 & \best 45.3  & \sbest 65.6 & \sbest 18.4 & 28.5        & \best 35.2  & 136.1        & 359.0       & 73.2        & 112.7        & 103.5       & 56.6        \\
    \bottomrule
    \end{tabular}
\end{table}

\subsection{Trainable parameters and inference latency}
Tab.~\ref{tab:supp:latency} reports the parameter counts and inference latency of the representation extractors used at inference for BAL and all compared methods. Inference latency is measured on a single NVIDIA A100 GPU with a batch size of 1, and may vary substantially depending on implementation details.

For a fairer comparison with methods that have fewer parameters than BAL, we additionally scale up these methods by increasing the number of layers to approximately match BAL's parameter count and evaluate the resulting models on each downstream task.
As shown in Tab.~\ref{tab:supp:paramFairComparison}, BAL still outperforms these methods on most metrics, showing that its gains are not simply due to model scale.

\begin{table}[t]
\centering
\caption{Comparison of parameter counts and inference latency for the representation extractors.}
\label{tab:supp:latency}
\vspace{5pt}
\footnotesize
\renewcommand{\arraystretch}{1.2}
\setlength{\tabcolsep}{1pt}
\begin{tabular}{@{}lcc@{}}
\toprule
\textbf{Method} & \textbf{\#Params [M]} & \textbf{Latency [ms]} \\
\midrule
A4Mer      & 9.64  & 16.0 \\
PUMPS      & 53.01 & 8.1  \\
USDRL      & 95.15 & 9.2  \\
BehaveMAE  & 5.81  & 4.3  \\
MotionBERT & 42.42 & 21.4 \\
MacDiff    & 11.96 & 5.3  \\
BAL        & 14.18 & 12.2 \\
\bottomrule
\end{tabular}
\end{table}

\begin{table*}[t!]
    \centering
    \fontsize{8pt}{9pt}\selectfont
    \caption{Parameter-matched comparison with scaled-up baselines. Methods that originally have fewer parameters than BAL are scaled to approximately match BAL's parameter count. BAL still outperforms them on most metrics.}
    \label{tab:supp:paramFairComparison}
    \renewcommand{\arraystretch}{1.2}
    \setlength{\tabcolsep}{1.8pt}
    \begin{tabular}{@{}l|c|cccccccc|ccc|ccc@{}}
    \toprule
    \multirow{3}{*}{\textbf{Method}}
    & \multirow{3}{*}{\textbf{\#Params [M]}}
    & \multicolumn{8}{c|}{\textbf{Captioning} $\uparrow$}
    & \multicolumn{3}{c|}{\multirow{2}{*}{\textbf{Forecasting} (mm) $\downarrow$}}
    & \multicolumn{3}{c}{\multirow{2}{*}{\textbf{Interpolation} (mm) $\downarrow$}} \\
    & & \multicolumn{2}{c}{BLEU-1}
      & \multicolumn{2}{c}{ROUGE-L}
      & \multicolumn{2}{c}{CIDEr}
      & \multicolumn{2}{c|}{BERTScore}
      & \multicolumn{3}{c|}{}
      & \multicolumn{3}{c}{} \\
    & & Nym & E3D
      & Nym & E3D
      & Nym & E3D
      & Nym & E3D
      & AMD & Nym & E3D
      & AMD & Nym & E3D \\
    \midrule
    A4Mer
    & 14.66
    & 6.4 & 54.2
    & 7.7 & 40.7
    & 2.1 & 12.4
    & 3.3 & 27.1
    & 235.2 & 313.4 & 136.0
    & 135.8 & 109.4 & 73.9 \\

    BehaveMAE
    & 14.19
    & 30.9 & \best 59.1
    & 37.8 & 44.6
    & 47.9 & 17.8
    & 26.0 & \sbest 34.1
    & 171.0 & \sbest 139.0 & 116.4
    & 139.1 & \sbest 98.3 & 76.9 \\

    MacDiff
    & 14.33
    & \best 39.0 & \best 59.1
    & \best 41.3 & \sbest 45.0
    & \sbest 62.4 & \sbest 18.5
    & \sbest 28.7 & \sbest 34.1
    & \sbest 142.9 & 367.1 & \sbest 84.5
    & \sbest 118.3 & 404.1 & \best 58.1 \\

    \textbf{BAL}
    & 14.18
    & \sbest 37.3 & \sbest 58.9
    & \sbest 41.2 & \best 45.3
    & \best 63.1 & \best 18.6
    & \best 29.6 & \best 35.0
    & \best 90.5 & \best 82.1 & \best 63.8
    & \best 117.5 & \best 77.0 & \sbest 59.2 \\
    \bottomrule
    \end{tabular}
\end{table*}

\section{Additional qualitative results}
\begin{figure}[t]
  \centering
  \includegraphics[width=\linewidth]{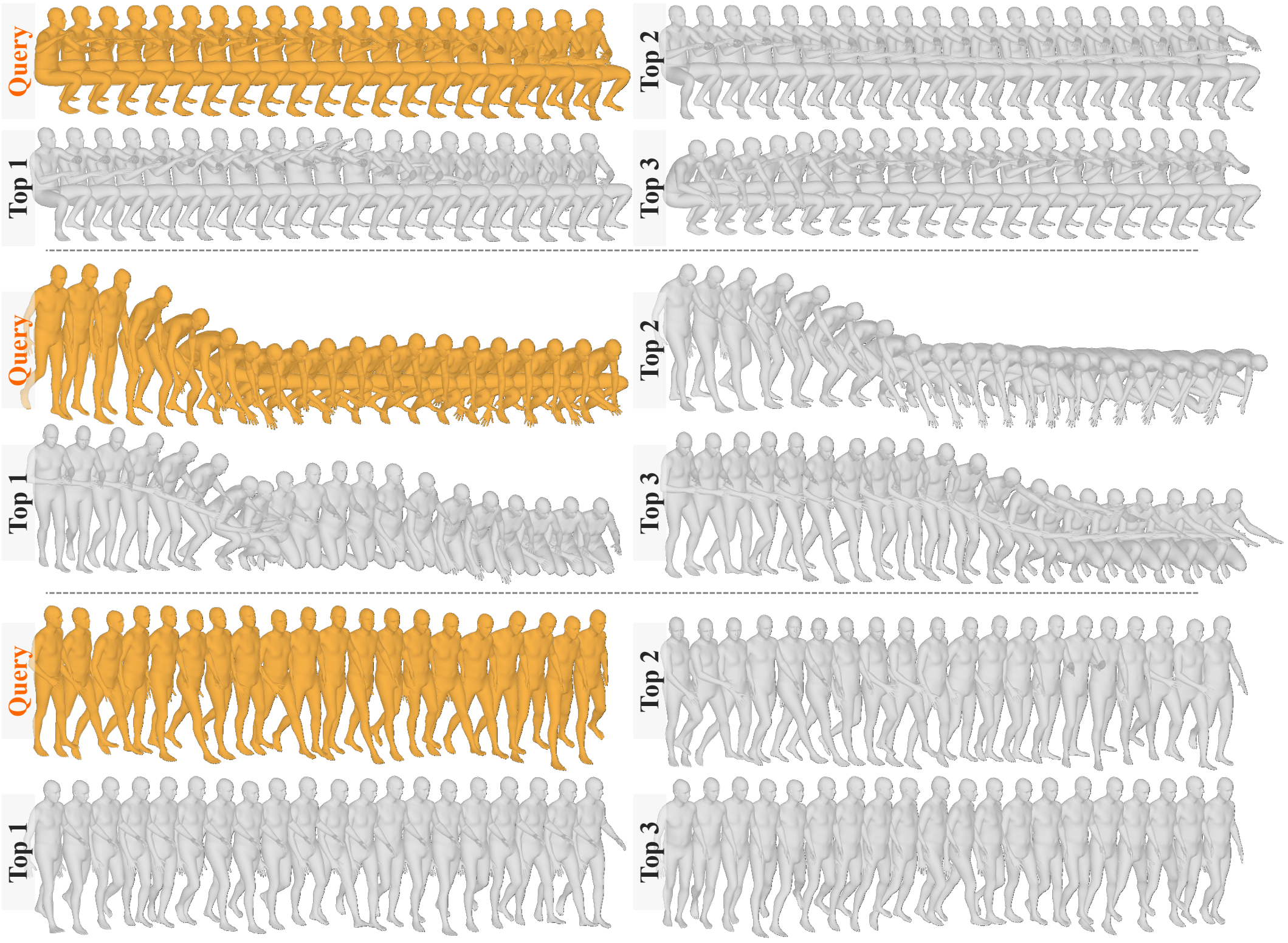}
  \caption{
    Motion retrieval examples. The nearest neighbors in the representation space correspond to semantically similar movements.
  }
  \label{fig:supp:retrieval1}
\end{figure}

\begin{figure}[t]
  \centering
  \includegraphics[width=\linewidth]{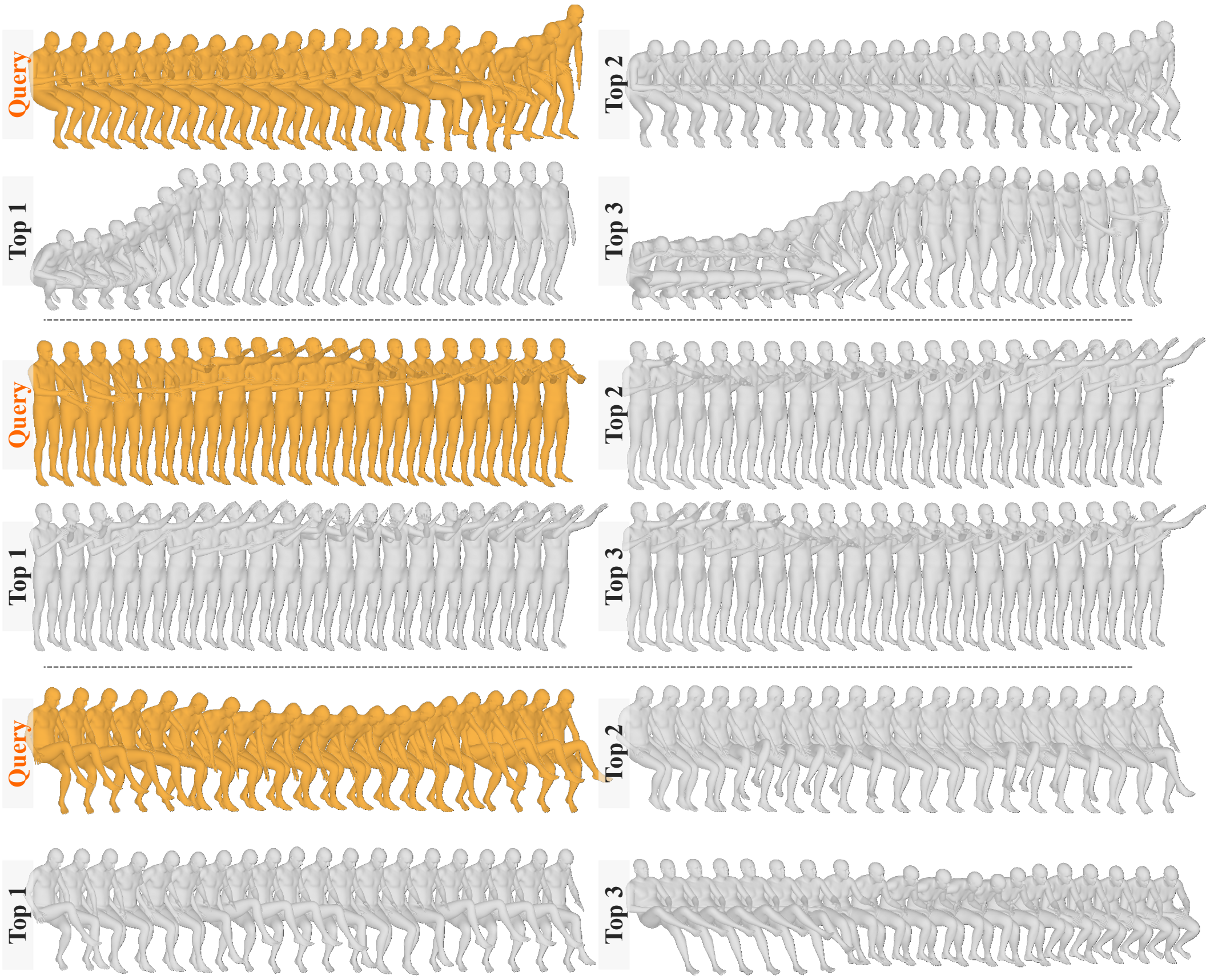}
  \caption{
    Additional motion retrieval examples.
  }
  \label{fig:supp:retrieval2}
\end{figure}

\begin{figure}[t]
  \centering
  \includegraphics[width=\linewidth]{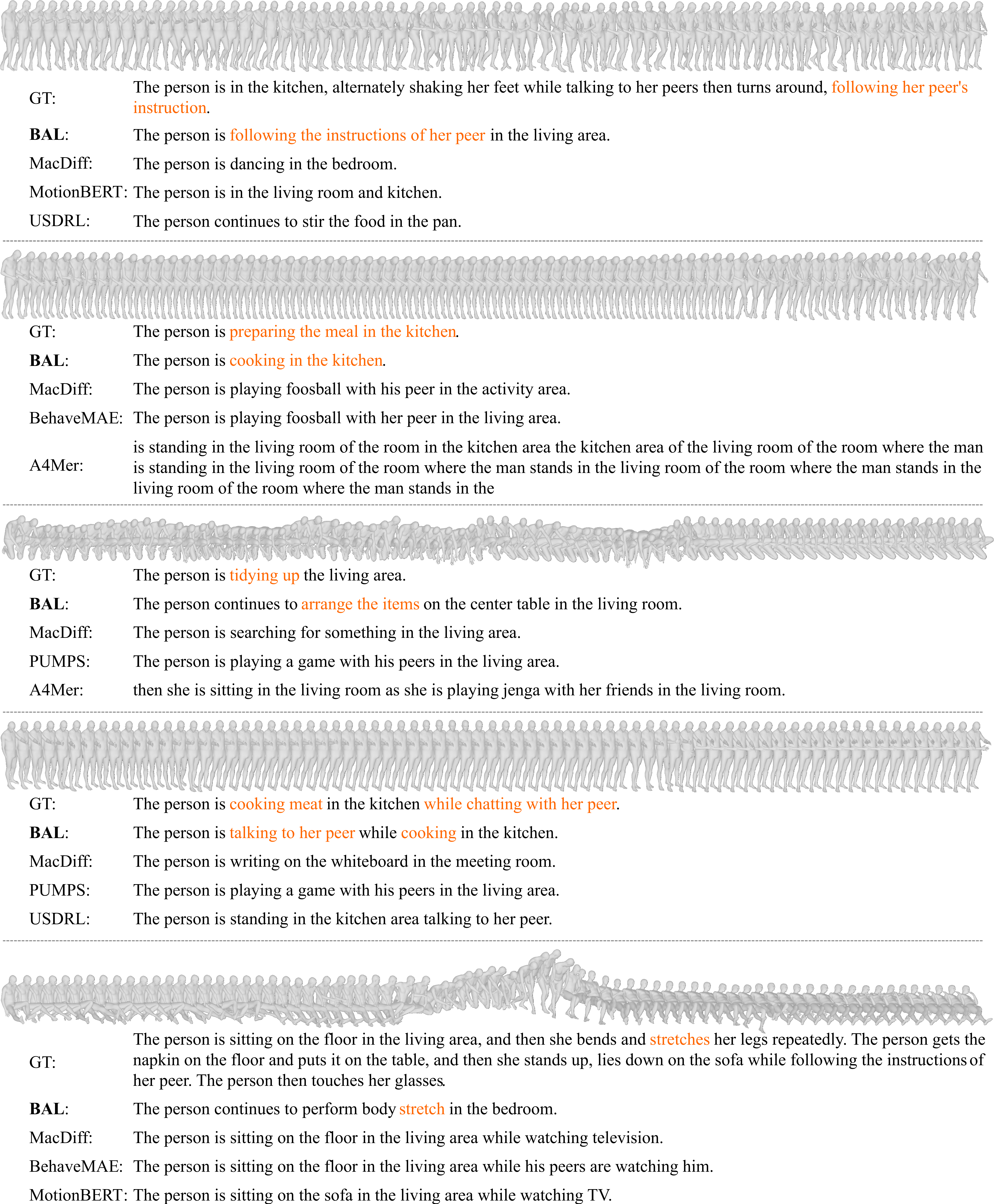}
  \caption{
    Qualitative results of motion captioning with the frozen LLM.
  }
  \label{fig:supp:captioning}
\end{figure}

\begin{figure}[t]
  \centering
  \includegraphics[width=\linewidth]{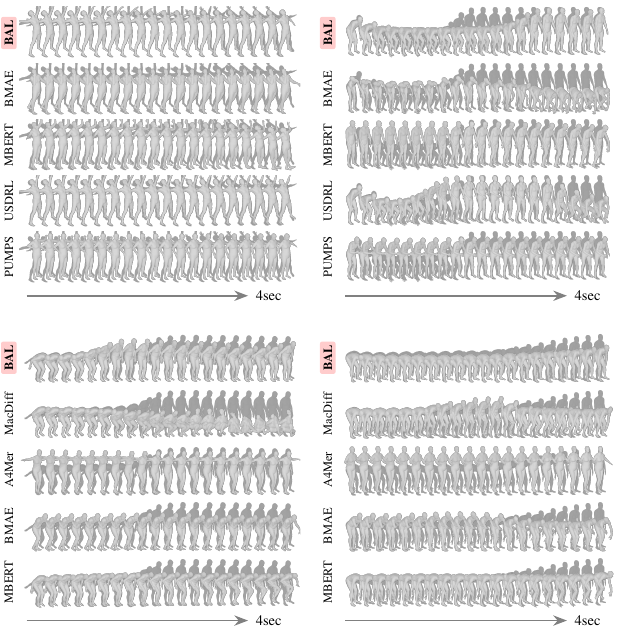}
  \caption{
    Forecasted poses via autoregressive latent token prediction and decoding. The SMPL meshes in dark color denote the ground truth.
  }
  \label{fig:supp:forecast}
\end{figure}

\begin{figure}[t]
  \centering
  \includegraphics[width=\linewidth]{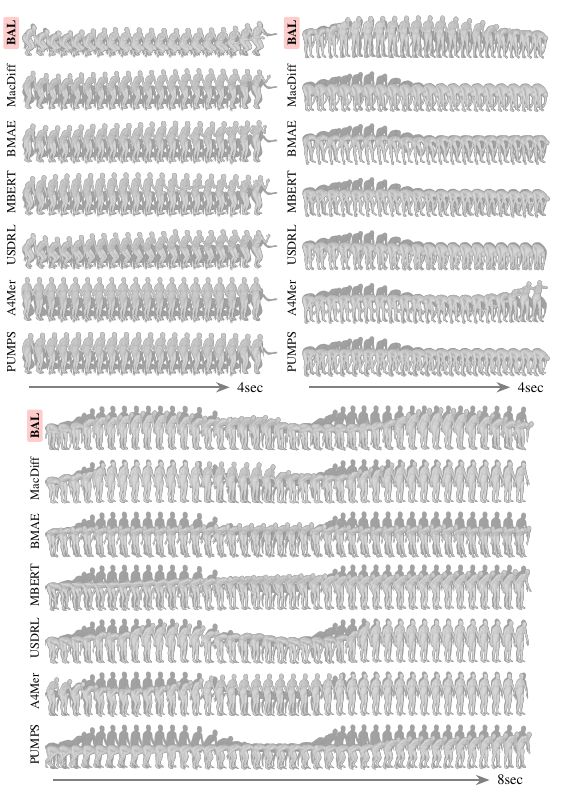}
  \caption{
    Interpolated poses via latent token interpolation and decoding. The dark-colored SMPL meshes are the ground truth.
  }
  \label{fig:supp:interp}
\end{figure}

We show additional qualitative results of motion retrieval (Figs.~\ref{fig:supp:retrieval1} and \ref{fig:supp:retrieval2}), captioning (Fig.~\ref{fig:supp:captioning}), forecasting (Fig.~\ref{fig:supp:forecast}), and interpolation (Fig.~\ref{fig:supp:interp}).


\clearpage
\section{Broader impact}
\label{sec:impact}
BAL provides a foundational framework for learning semantic representations of human movement from 3D pose sequences, contributing to advances in human behavior modeling.
These representations may support applications involving semantic understanding of human movement, such as human-computer interaction, assistive technologies, rehabilitation analysis, and robotics.
At the same time, improved movement representations could be misused for unauthorized surveillance or privacy-invasive behavioral profiling, especially if applied to data collected without consent or adequate safeguards.
Although this work focuses on 3D pose sequences, similar principles may also be explored in other sequential modalities.
Applications and extensions of this work should be developed with careful attention to consent, data privacy, fairness, and domain-specific ethical guidelines.

\end{document}